%% file: main.tex
\documentclass{article}
\usepackage{iclr2027_conference,times}
\input{math_commands.tex}

\usepackage{hyperref}
\usepackage{url}
\usepackage{graphicx}
\usepackage{booktabs}
\usepackage{multirow}
\usepackage{amsmath,amssymb}
\usepackage{xcolor}
\usepackage{caption}
\usepackage{subcaption}
\usepackage{enumitem}
\graphicspath{{figures/}}
\newcommand{\shortpar}{\looseness=-1\relax}

\iclrfinalcopy

\title{Language as the Interface: Foundation-Model Contrastive Learning Links Transcriptomes and Electrophysiology}

\author{Junbo Shen$^{1}$ \quad Jinying Gao$^{2,3,4}$ \quad Bo Lei$^{3}$\thanks{Corresponding author: Bo Lei (\texttt{leibo@baai.ac.cn}).} \\
$^{1}$Department of Computer Science and Engineering, The Chinese University of Hong Kong \\
$^{2}$State Key Laboratory of Brain Cognition and Brain-inspired Intelligence Technology, \\
\hphantom{$^{2}$}Institute of Automation, Chinese Academy of Sciences, Beijing, China \\
$^{3}$Beijing Academy of Artificial Intelligence, Beijing, China \\
$^{4}$University of Chinese Academy of Sciences, Beijing, China}

\newcommand{\method}{\textsc{LangPatch}}

\begin{document}
\maketitle
\lhead{Preprint.}

\begin{abstract}
Integrating transcriptomic and electrophysiological data is essential for building multimodal foundation models for neuroscience. Patch-seq provides paired measurements of gene expression and intrinsic electrophysiology from the same neuron, establishing a basis for training cross-modal models. Here we introduce \method{}, a foundation-model-based contrastive learning framework that uses paired Patch-seq data to align pretrained GenePT representations with electrophysiological phenotypes through a language-based interface. Gene descriptions and verbalized electrophysiological profiles are embedded by the same frozen text encoder. A context adapter and projection modules connect the modalities through paired contrastive learning. Across mouse visual, mouse motor, and human cortical cohorts, \method{} achieves the highest mean transcriptome-to-electrophysiology prediction correlation among the evaluated foundation-model and representation-learning methods. It also improves held-out cross-modal alignment in the two mouse cohorts (FOSCTTM 0.107 in visual cortex and 0.135 in motor cortex, against 0.208 and 0.222 for JAMIE, an existing cross-modal Patch-seq imputation method). It predicts transcriptomic family, type, cortical layer, and marker-gene expression from electrophysiology, exceeding other baselines on most endpoints. More importantly, the method transfers across brain areas, laboratories and species: a model trained on mouse visual cortex predicts electrophysiology in motor cortex with approximately 70\% correlation retention and in human cortex with 47\% (58\% on acute-slice recordings); a model trained on human cortex predicts mouse electrophysiology with 63\% correlation retention, and the method applies to another laboratory's electrophysiological features without retraining. Together, these results demonstrate alignment between molecular and functional representations of neurons, providing a building block for multimodal foundation models in neuroscience.
\end{abstract}

\input{1_intro}

\input{2_method}

\input{3_experiments_results_discussions}

\input{4_statements}

\bibliography{refs}
\bibliographystyle{iclr2027_conference}

\clearpage
\appendix
\input{appendix}

\end{document}

%% file: math_commands.tex
\usepackage{amsmath,amsfonts,bm}

\def\eqref#1{equation~\ref{#1}}
\def\1{\bm{1}}

\def\vc{{\bm{c}}}

\def\ve{{\bm{e}}}

\def\vg{{\bm{g}}}
\def\vh{{\bm{h}}}

\def\vw{{\bm{w}}}
\def\vx{{\bm{x}}}
\def\vy{{\bm{y}}}
\def\vz{{\bm{z}}}

\def\mG{{\bm{G}}}

\def\mI{{\bm{I}}}

\def\mW{{\bm{W}}}

\def\mZ{{\bm{Z}}}

\DeclareMathAlphabet{\mathsfit}{\encodingdefault}{\sfdefault}{m}{sl}
\SetMathAlphabet{\mathsfit}{bold}{\encodingdefault}{\sfdefault}{bx}{n}

\newcommand{\R}{\mathbb{R}}

%% file: 1_intro.tex
\section{Introduction}
\label{sec:intro}
Joint modeling of molecular processes, neural activity, and behavior may be central to building neuroscience foundation models capable of capturing the complexity of the brain~\citep{arkhipov2025integrating,wang2025foundation,mathis2026joint}. Recent work has advanced neural activity prediction~\citep{wang2025foundation} and joint neural--behavioral modeling~\citep{schneider2023learnable,willeke2026omnimouse}, providing tools for basic neuroscience research. While foundation models for single-cell transcriptomics have advanced substantially~\citep{theodoris2023transfer,cui2024scgpt,hao2024scfoundation,schaar2024nicheformer}, joint modeling of transcriptomic and functional data remains comparatively limited within pretrained frameworks, despite existing approaches for cross-modal alignment~\citep{gala2021consistent,cohenkalafut2023jamie}.

Patch-seq data provide a valuable empirical testbed for developing models that align molecular and functional representations of neurons. By pairing transcriptomic profiles with intrinsic electrophysiological measurements from the same neuron, these data support both cross-modal learning and evaluation~\citep{cadwell2016electrophysiological,fuzik2016integration}. In parallel, advances in pretrained gene representations offer an opportunity to incorporate knowledge acquired beyond individual Patch-seq cohorts~\citep{theodoris2023transfer,cui2024scgpt,hao2024scfoundation,schaar2024nicheformer,chen2024genept}. Currently, the key question is how to use these paired measurements to align pretrained molecular representations with electrophysiological phenotypes.

Maps between the two modalities matter beyond the cohort that produced them: a forward model can impute physiology for the millions of cells in transcriptomic atlases that will never be patched, and a reverse model assigns type, layer and marker expression to a recorded cell without sequencing it.\shortpar

Multimodal single-cell neuroscience is a low-resource regime. A Patch-seq cohort holds $10^3$--$4\cdot10^3$ neurons, each modality has its own feature vocabulary, and laboratories extract different features with different pipelines. The prevailing approach---pretrain a single-cell foundation model on tens of millions of cells, then adapt its weights to the target task~\citep{theodoris2023transfer,rosen2023uce,schaar2024nicheformer,cui2024scgpt}---assumes access to those weights and enough cells to move hundreds of millions of parameters. On these cohorts frozen embeddings from three such models are at best on par with a frozen text prior, and on the 1,208-cell motor cortex cohort one of them loses most of its accuracy (Sec.~\ref{sec:results-forward}), consistent with prior evaluations~\citep{kedzierska2025zeroshot,ahlmann2025deep,wenteler2024pca}. Cross-modal Patch-seq methods instead train a pair of encoders from scratch on the cohort itself~\citep{gala2019coupled,gala2021consistent,cohenkalafut2023jamie}, so the representation starts from no knowledge about genes, and nothing learned on one cohort carries over to the next feature vocabulary.

We take a different route and use language as the interface between the modalities (Fig.~\ref{fig:method}). Both are verbalizable: a gene has a textual description of its function, and an electrophysiological feature vector is a short list of named quantities with units. We embed both with the same frozen general-purpose text embedder, which thereby serves as the shared embedding space of two measured modalities, neither of which is text. The transcriptomic side enters through GenePT embeddings of gene descriptions~\citep{chen2024genept}, corrected by a 0.4M-parameter adapter conditioned on per-gene statistics of the training cohort; the electrophysiological side enters through a deterministic verbalization of the feature record. No modality-specific encoder is trained from scratch: two linear projections trained with a contrastive objective~\citep{oord2018representation,radford2021learning} suffice to align the two modalities, and the embedder's weights are never updated. Verbalization also makes the electrophysiological side independent of a fixed feature set: a cohort with different features is verbalized with its own feature names and embedded by the same model.

We call the method \method{} and evaluate it on the visual cortex cohort of \citet{gouwens2020integrated} (3,654 cells), the motor cortex cohort of \citet{scala2021phenotypic} (1,208) and the human cortex cohort of \citet{lee2023signature} (704 paired cells; 612 in the in-domain benchmark). Our contributions are:
\begin{itemize}[leftmargin=*,itemsep=1pt]
\item \textbf{A frozen language model as the shared embedding space.} To our knowledge, \method{} is the first method that uses a foundation model to link a neuron's gene expression with its electrophysiology. A transcriptome is represented through the text descriptions of its genes and a recording through a verbalized list of its measurements, both embedded by a single frozen text embedder; alignment requires only a 0.4M-parameter adapter and two linear projections, trained in about one minute on one GPU, and updates no foundation-model weight (Sec.~\ref{sec:method}).
\item \textbf{Accuracy among learned methods.} \method{} is the most accurate transcriptomic-to-electrophysiology features predictor among foundation-model and representation-learning methods on two mouse cortices and a human cortex, where single-cell foundation models are at best on par with a frozen text prior; it gives the best held-out cross-modal alignment (FOSCTTM 0.107 in visual cortex and 0.135 in motor cortex, against 0.208 and 0.222 for JAMIE), and reversely, electrophysiology features-to-transcriptomic predictions (family macro-F1 0.78 and 0.71, marker $r$ 0.54 and 0.57) exceed per-task supervised heads, embedding baselines and JAMIE (Secs.~\ref{sec:results-forward}, \ref{sec:results-reverse}).
\item \textbf{Zero-shot transfer across areas, laboratories and species.} Trained in different cortical areas of the mouse, the method retains 70\% and 71\% of its in-domain correlation across cortical areas, with all 11 shared features predicted better than chance, and from mouse to human cortex, 18 of 25 electrophysiological features transfer better than chance, with 47\% of the in-domain correlation retained overall and 58\% on acute slices; 20 paired cells recover most of the calibration gain. From human to mouse, 63\% of the in-domain correlation is retained, and the method applies to another laboratory's electrophysiological features without retraining, which is infeasible for encoders with numerical input (Sec.~\ref{sec:results-transfer}).
\item \textbf{Necessity of the language interface.} Replacing the text embeddings of both modalities with numerical encoders lowers the prediction correlation from 0.521 to 0.419 in visual cortex and from 0.532 to 0.495 in motor cortex, alignment improves with the quality of the text embedder, and the full method outperforms each of its ingredients used alone (Sec.~\ref{sec:results-ablation}).
\item \textbf{An interpretable shared space.} The aligned space separates transcriptomic families for both modalities, and, because the representation is linear in expression, each electrophysiological feature is attributed to named genes; the prioritized genes recover cell-type markers and, more weakly, ion-channel genes (Sec.~\ref{sec:discussion}).
\end{itemize}

\section{Related Work}
\label{sec:related}

\paragraph{Cross-modal learning on Patch-seq.}
Statistical models first related transcriptomic types and single genes to electrophysiology~\citep{tripathy2017transcriptomic,bomkamp2019transcriptomic,kobak2021sparse}. Coupled autoencoders~\citep{gala2019coupled,gala2021consistent} train one encoder per modality on the cohort and tie their latent spaces, and deepManReg~\citep{nguyen2021deepmanreg} aligns modalities through manifold regularization. JAMIE~\citep{cohenkalafut2023jamie} learns a joint variational latent with cross-modal imputation and is the strongest published baseline; MMIDAS~\citep{bahrami2024mmidas} separates discrete and continuous factors of variation, and \citet{schwider2026cross} map human interneuron electrophysiology to transcriptomic subclass with numeric features and mouse pretraining (macro-F1 0.66 to 0.68 on Lee~2023). All of these methods train a separate encoder per modality on the numerical features of each cohort; \method{} instead embeds both modalities with one pretrained text embedder.

\paragraph{Language representations of genes and cells.}
GenePT~\citep{chen2024genept} embeds textual gene descriptions with a general-purpose text embedder~\citep{openai2024embeddings} and represents a cell as the expression-weighted average of its genes; scELMo~\citep{liu2023scelmo} extends the idea to metadata. Cell2Sentence~\citep{levine2023cell2sentence} renders a cell as a rank-ordered gene sentence for a language model, and CellWhisperer~\citep{schaefer2025cellwhisperer} aligns transcriptomes with free-text annotations contrastively. Each of these models relates a single measured modality, the transcriptome, to text; none includes electrophysiology or another measured modality. TabLLM~\citep{hegselmann2023tabllm} serializes numeric table rows into text for few-shot classification, the closest precedent for our verbalized records; we serialize to place a second modality in the same embedding space as the first.

\paragraph{Contrastive alignment and adaptation.}
Our alignment objective is the symmetric InfoNCE loss of CLIP~\citep{oord2018representation,radford2021learning}, which scCLIP~\citep{xiong2023scclip} and scPairing~\citep{scpairing2025} apply to paired multi-omic single-cell measurements; both train modality-specific encoders, whereas our towers share one frozen embedder and are joined by linear maps. Parameter-efficient adaptation with adapters~\citep{houlsby2019parameter} or low-rank updates~\citep{hu2022lora} modifies a network from the inside and requires its weights; feature adapters such as CLIP-Adapter and Tip-Adapter~\citep{gao2024clipadapter,zhang2022tipadapter} add a residual on frozen encoder outputs, as we do, but ours is conditioned on cohort statistics rather than on the sample and bounded by a gate. In contrast to these lines of work, \method{} uses a single frozen general-purpose text embedder as the shared embedding space of two measured modalities, neither of which is text, so that no modality-specific encoder is trained from scratch. To our knowledge, this is the first such approach for single-cell data, and Sec.~\ref{sec:results-ablation} shows that it outperforms each of its ingredients used alone.

%% file: 2_method.tex
\section{Method}
\label{sec:method}

\paragraph{Setting.}
A Patch-seq cohort provides, for each neuron $i=1,\dots,N$, a transcriptome $\vx_i\in\R^{G}$ over a gene panel of size $G$ and a vector of intrinsic electrophysiological features $\vy_i\in\R^{F}$ extracted from the same cell's recordings. We use the log-normalized expression $x_{ij}=\log(1+10^4\,c_{ij}/\sum_{j'}c_{ij'})$ of the raw counts $c_{ij}$ over the panel. Cohorts are small ($N\approx10^3$--$4\cdot10^3$), the feature vocabulary of $\vy$ differs between laboratories, and labels exist for the transcriptomic side only. We want one representation that supports $\vx\to\vy$, the reverse readout $\vy\to$ transcriptomic identity, and transfer of both to new cohorts.

\begin{figure}[t]\centering\includegraphics[width=\linewidth]{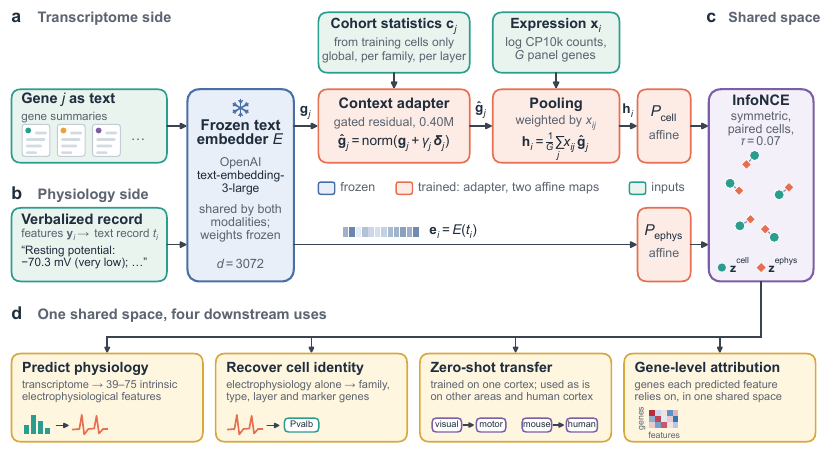}\caption{Language as the interface. One frozen text embedder $E$ (text-embedding-3-large) embeds both modalities. (a) Gene descriptions are embedded by $E$ (these are the GenePT embeddings); a 0.4M-parameter adapter conditioned on training-cohort statistics adjusts the frozen gene prior, and a cell is the expression-weighted mean of its adapted genes followed by a linear map. (b) Electrophysiology is verbalized and embedded by the same embedder $E$, then linearly mapped. (c) A symmetric InfoNCE objective on paired cells aligns the two; the embedder is frozen. (d) The shared space has numerous downstream applications: predicting electrophysiological features from the transcriptome (Sec.~\ref{sec:results-forward}), reading transcriptomic identity from electrophysiology alone (Sec.~\ref{sec:results-reverse}), zero-shot transfer across cortical areas, laboratories and species (Sec.~\ref{sec:results-transfer}), and gene-level attribution of each feature (Sec.~\ref{sec:discussion}).}\label{fig:method}\end{figure}

\paragraph{Overview (Fig.~\ref{fig:method}).}
Both modalities are expressed as text and encoded by the \emph{same frozen} general-purpose text embedder $E(\cdot)\in\R^{d}$ ($d=3072$; text-embedding-3-large). A gene is represented by the embedding of its functional description (the GenePT embeddings of \citealp{chen2024genept}, which are produced by this embedder), and a recording by the embedding of a deterministic verbalization of its feature vector. The only trained representation-stage components are a small \emph{context adapter} that conditions the frozen gene prior on per-gene statistics of the training cohort (global, and per transcriptomic family and layer of the training cells), and two affine projections trained with a contrastive objective. No embedder weights are updated, and the embedder is used as a black box. The text embedder is thus the shared embedding space of both modalities, and no modality-specific encoder is trained from scratch.

\subsection{Gene side: a frozen language prior, adapted by dataset statistics}
\label{sec:method-gene}

Gene $j$ is represented by $\vg_j=E(\text{description of gene }j)$, computed with the same embedder $E$ as the recordings. These vectors are the released GenePT embeddings~\citep{chen2024genept}, which are text-embedding-3-large applied to each gene's NCBI and UniProt summary. For each gene we compute a \emph{context vector} $\vc_j\in\R^{C}$ from the training cells only: four global statistics of $x_{\cdot j}$ (mean, standard deviation, detection rate, coefficient of variation) and, for each transcriptomic family and cortical layer present in the training set, the family- and layer-conditional mean and detection rate ($C=24$ for mouse visual cortex, 30 for motor cortex, 18 for human cortex; Appendix~\ref{app:hyper}). The adapter maps the context to a gated residual on the prior,
\begin{equation}
\bm{\delta}_j=\mW_2\,\mathrm{GELU}\big(\mW_1\,\mathrm{LN}(\vc_j)\big),\qquad
\gamma_j=s\,\sigma\big(\vw^{\top}\mathrm{LN}(\vc_j)\big),\qquad
\hat{\vg}_j=\frac{\vg_j+\gamma_j\bm{\delta}_j}{\lVert\vg_j+\gamma_j\bm{\delta}_j\rVert_2},
\label{eq:adapter}
\end{equation}
with $\mW_1\in\R^{128\times C}$, $\mW_2\in\R^{d\times128}$, a scalar gate bounded by $s=0.1$, and $\mathrm{LN}$ layer normalization. The adapter has $0.40$M parameters; it can only move each gene a bounded distance from its prior, and a cosine penalty $\lambda_g\,(1-\cos(\hat{\vg}_j,\vg_j))$ keeps it close. A cell is the expression-weighted mean of its adapted gene vectors, followed by a linear projection,
\begin{equation}
\vh_i=\frac{1}{G}\sum_{j=1}^{G}x_{ij}\,\hat{\vg}_j,\qquad
\vz^{\mathrm{cell}}_i=\mathrm{normalize}\big(P_{\mathrm{cell}}(\vh_i)\big),\qquad
P_{\mathrm{cell}}=\mathrm{LN}\circ\mathrm{Linear}_{d\to d}\circ\mathrm{LN}.
\label{eq:cell}
\end{equation}
Because Eq.~(\ref{eq:cell}) is linear in $\vx_i$ once the adapted gene matrix $\hat{\mG}=[\hat{\vg}_1;\dots;\hat{\vg}_G]$ is fixed, the representation is a fixed linear map of expression whose geometry is inherited from the language prior; Sec.~\ref{sec:discussion} uses this linearity to attribute predictions to genes.

\subsection{Electrophysiology side: verbalize, then embed}
\label{sec:method-ephys}

Each feature vector is rendered as a short record (``\emph{Cell electrophysiology profile. Resting membrane potential [mV]: $-70.3$ (very low); \dots}'') and embedded once by the same frozen embedder.
Feature names are expanded to readable names with units, values are winsorized to the 1st--99th percentiles, and each value is followed by its within-cohort quintile label; four coarse profile tokens summarize excitability, adaptation, spike shape and membrane state (Appendix~\ref{app:text}). The record $t_i$ is embedded as $\ve_i=E(t_i)$ and projected, $\vz^{\mathrm{ephys}}_i=\mathrm{normalize}(P_{\mathrm{ephys}}(\ve_i))$, with $P_{\mathrm{ephys}}$ of the same form as $P_{\mathrm{cell}}$. Verbalization makes the electrophysiology side vocabulary-agnostic in principle: a cohort with a different feature set produces a record in the same language and lands in the same embedding space; Sec.~\ref{sec:results-transfer} tests this on two cohorts whose feature sets share no names.

\subsection{Alignment}
\label{sec:method-align}

The two projections and the adapter are trained jointly on paired cells with a symmetric InfoNCE objective~\citep{oord2018representation,radford2021learning} at temperature $\tau=0.07$,
\begin{equation}
\mathcal{L}_{\mathrm{align}}=\tfrac12\big[\mathrm{CE}(\mZ^{\mathrm{ephys}}\mZ^{\mathrm{cell}\top}/\tau,\,\mI)+\mathrm{CE}(\mZ^{\mathrm{cell}}\mZ^{\mathrm{ephys}\top}/\tau,\,\mI)\big],
\label{eq:clip}
\end{equation}
plus three light regularizers: an auxiliary regression of the standardized features $\vy_i$ from both towers (which anchors the shared space to measurable quantities), the gene-cosine penalty above ($\lambda_g=0.5$), and a cosine penalty that keeps $\vh_i$ close to the unadapted cell embedding ($\lambda_c=0.1$), with an $\ell_2$ penalty on the residual ($10^{-4}$). Training is full-batch AdamW for 2{,}000 epochs with a warm-up in which only the auxiliary regression is active (Appendix~\ref{app:hyper}); the checkpoint with the lowest validation contrastive loss is kept. Sec.~\ref{sec:results-ablation} ablates the adapter, the two cosine penalties and the auxiliary regression. Projections are affine maps between layer normalizations, kept linear because this sufficed empirically and keeps the representation linear in expression.

\subsection{Readouts}
\label{sec:method-readout}

\textbf{Forward.} Each electrophysiological feature $f$ gets its own small regressor $\phi_f(\vz^{\mathrm{cell}}_i)$ (an MLP whose width is selected on the validation fold among four candidates: 512--128 units on both mouse cohorts, 256 on human cortex; Appendix~\ref{app:hyper}), trained on the standardized target with a log transform for heavy-tailed positive features; the epoch is selected per feature on the validation fold.
\textbf{Reverse.} A regressor maps the standardized feature vector $\vy_i$ into the aligned transcriptomic space; the predicted embedding, compressed to its leading directions, is concatenated with $\vy_i$ and read by one multitask head that predicts the transcriptomic family, the fine type, the cortical layer, the log-expression of 27 marker genes and 9 marker programs (Appendix~\ref{app:reverse}). All readout heads are fit on training cells and evaluated on held-out cells of the same fold.
\textbf{Transfer.} The adapted gene matrix $\hat{\mG}$, the projections and the heads are used unchanged on a new cohort: its expression is normalized over the source panel and passed through Eqs.~(\ref{eq:adapter})--(\ref{eq:cell}). Feature correspondences between laboratories are needed only for \emph{scoring}; they are fixed before any target score is computed (Appendix~\ref{app:crosswalk}). The embedder is never updated; parameter counts are listed in Appendix~\ref{app:hyper}. A linear readout on $\vz^{\mathrm{cell}}$ matches or exceeds the MLP heads on the human cohort (Appendix~\ref{app:human-indomain}), so the nonlinear heads are a refinement rather than part of the representation.

%% file: 3_experiments_results_discussions.tex
\section{Experimental setup}
\label{sec:setup}

\paragraph{Cohorts (Table~\ref{tab:datasets} in Appendix~\ref{app:hyper}).}
Three Patch-seq cohorts serve as in-domain benchmarks: mouse visual cortex~\citep{gouwens2020integrated}, 3{,}654 GABAergic interneurons with 39 IPFX features~\citep{allen2024ipfx} over a 1{,}293-gene panel; mouse motor cortex~\citep{scala2021phenotypic}, 1{,}208 excitatory and inhibitory neurons with 29 features over 1{,}277 genes; and human cortex~\citep{lee2023signature}, 612 GABAergic interneurons with complete 75-feature profiles, mapped onto the visual panel by upper-cased symbol identity (100\%; not a curated orthology). %
The zero-shot target of the visual-cortex model is the set of all Lee~2023 cells with electrophysiology (704; 247 from acute and 457 from cultured slices). %

\paragraph{Protocol.}
Each cohort is split into five type-stratified folds (train/validation/test $\approx$ 65/15/20; seed 0); all fitting and selection use the fold's training and validation cells only, and we report means over the five test folds. %
Folds partition one cohort and are not replicates, so significance is assessed with paired Wilcoxon tests and win counts (binomial tests) over features.
Forward accuracy is the per-feature Pearson $r$ on held-out cells (mean over features, then folds), plus AUROC for above/below-median and raw MAE; alignment of the shared space is scored on held-out cells by FOSCTTM, the fraction of cells of the other modality that lie closer to a cell than its true partner (0 perfect, 0.5 chance), and by label-transfer accuracy (LTA), the accuracy of a $k$-nearest-neighbour classifier that carries transcriptomic labels from the electrophysiology embeddings to the transcriptome embeddings of the same cells (Appendix~\ref{app:integration}); reverse readouts use macro-F1 for family, fine type and layer, and Pearson $r$ and AUPRC for marker genes and programs (Appendix~\ref{app:reverse}).

\paragraph{Baselines.}
Forward: a \emph{Plain GenePT MLP} (frozen pooled GenePT cell embedding and one multi-output MLP: our method without adapter and alignment), MLPs on frozen Geneformer~\citep{theodoris2023transfer}, UCE~\citep{rosen2023uce} and Nicheformer~\citep{schaar2024nicheformer} embeddings, Nicheformer with its last transformer block trained end-to-end on the cohort together with the head, a GenePT--ephys VAE, and JAMIE~\citep{cohenkalafut2023jamie} re-run on the same folds.
Integration: JAMIE, the VAE, and Nicheformer embeddings aligned with the embedded records by CLIP-style contrastive learning (\emph{Nicheformer+CLIP}).
Reverse: per-task supervised heads (logistic regression, ridge), regressors from electrophysiology into gene counts, GenePT (PCA-ridge, PLS, MLP) and scFM embeddings, the VAE, and JAMIE's reverse map.

\paragraph{Transfer scoring.}
Because electrophysiological feature vocabularies differ across cohorts, transfer is scored on matched biophysical \emph{concepts}: 11 between mouse visual and motor cortex (e.g., input resistance, rheobase, spike amplitude; Table~\ref{tab:app-crosswalk-mouse}) and 25 for Lee~2023, with one-to-one feature matches and expected signs fixed before scoring (Appendix~\ref{app:crosswalk}). A concept passes if the Pearson $r$ between the five-fold ensemble prediction and the target measurement exceeds the 95th percentile of 1{,}000 cell-permutation nulls with the expected sign. \emph{Retention} $R$ is the mean target $|r|$ divided by the in-domain held-out mean $|r|$ over the same concepts. Pre-registered verdicts are A if $R\ge0.5$ and at least two thirds of concepts pass, B if $R\ge0.3$, and C otherwise. Human-to-mouse transfer is scored on the same concepts in the reverse direction (Appendix~\ref{app:ablations}). Absolute scale is recovered by affine calibration on $n$ paired target cells (200 resamples).

\section{Results}
\label{sec:results}

\subsection{Forward prediction}
\label{sec:results-forward}

\suppressfloats[t] %
\begin{table}[t]\centering\caption{Forward prediction on held-out cells (mean $\pm$ s.d.\ over five folds): per-feature Pearson $r$, AUROC and raw MAE. Ours: context-adapted GenePT with alignment and per-feature heads whose width is selected on the validation fold (Appendix~\ref{app:hyper}).}\label{tab:forward}
\small
\setlength{\tabcolsep}{3pt}
\resizebox{\textwidth}{!}{%
\begin{tabular}{l ccc ccc ccc}
\toprule
 & \multicolumn{3}{c}{Mouse visual (39 feat.)} & \multicolumn{3}{c}{Mouse motor (29 feat.)} & \multicolumn{3}{c}{Human Lee (75 feat.)} \tabularnewline
\cmidrule(lr){2-4}\cmidrule(lr){5-7}\cmidrule(lr){8-10}
Method & $r\uparrow$ & AUROC$\uparrow$ & MAE$\downarrow$ & $r\uparrow$ & AUROC$\uparrow$ & MAE$\downarrow$ & $r\uparrow$ & AUROC$\uparrow$ & MAE$\downarrow$ \tabularnewline
\midrule
Ours & \textbf{0.522 $\pm$ 0.018} & \textbf{0.779 $\pm$ 0.008} & \textbf{8.98 $\pm$ 0.12} & \textbf{0.520 $\pm$ 0.021} & \textbf{0.801 $\pm$ 0.005} & \textbf{7.72 $\pm$ 0.33} & \textbf{0.414 $\pm$ 0.030} & \textbf{0.725 $\pm$ 0.014} & \textbf{9.82 $\pm$ 0.27} \tabularnewline
\midrule
Plain GenePT MLP & 0.505 $\pm$ 0.016 & 0.768 $\pm$ 0.005 & 9.22 $\pm$ 0.04 & 0.502 $\pm$ 0.020 & 0.790 $\pm$ 0.002 & 8.02 $\pm$ 0.37 & 0.387 $\pm$ 0.021 & 0.713 $\pm$ 0.009 & 9.90 $\pm$ 0.19 \tabularnewline
Geneformer MLP & 0.492 $\pm$ 0.017 & 0.760 $\pm$ 0.005 & 9.54 $\pm$ 0.16 & 0.146 $\pm$ 0.028 & 0.574 $\pm$ 0.010 & 10.44 $\pm$ 0.32 & 0.401 $\pm$ 0.034 & 0.719 $\pm$ 0.010 & 9.88 $\pm$ 0.27 \tabularnewline
UCE MLP & 0.410 $\pm$ 0.016 & 0.714 $\pm$ 0.007 & 10.51 $\pm$ 0.14 & 0.462 $\pm$ 0.014 & 0.762 $\pm$ 0.009 & 8.57 $\pm$ 0.38 & 0.312 $\pm$ 0.035 & 0.670 $\pm$ 0.018 & 10.65 $\pm$ 0.37 \tabularnewline
Nicheformer MLP & 0.486 $\pm$ 0.016 & 0.756 $\pm$ 0.005 & 9.56 $\pm$ 0.08 & 0.503 $\pm$ 0.018 & 0.784 $\pm$ 0.007 & 8.06 $\pm$ 0.27 & 0.386 $\pm$ 0.028 & 0.710 $\pm$ 0.008 & 10.13 $\pm$ 0.24 \tabularnewline
Fine-tuned Nicheformer & 0.490 $\pm$ 0.017 & 0.760 $\pm$ 0.007 & 9.47 $\pm$ 0.19 & 0.510 $\pm$ 0.013 & 0.787 $\pm$ 0.004 & 8.04 $\pm$ 0.36 & 0.386 $\pm$ 0.023 & 0.710 $\pm$ 0.008 & 10.06 $\pm$ 0.22 \tabularnewline
GenePT--ephys VAE & 0.455 $\pm$ 0.047 & 0.734 $\pm$ 0.028 & 9.88 $\pm$ 0.36 & 0.021 $\pm$ 0.019 & 0.513 $\pm$ 0.015 & 10.72 $\pm$ 0.33 & 0.001 $\pm$ 0.018 & 0.510 $\pm$ 0.010 & 11.82 $\pm$ 0.15 \tabularnewline
JAMIE & 0.453 $\pm$ 0.016 & 0.739 $\pm$ 0.005 & 10.05 $\pm$ 0.20 & 0.432 $\pm$ 0.027 & 0.757 $\pm$ 0.007 & 8.91 $\pm$ 0.30 & 0.406 $\pm$ 0.034 & 0.718 $\pm$ 0.010 & 10.03 $\pm$ 0.28 \tabularnewline
\bottomrule
\end{tabular}
}
\end{table}

Among the methods compared, \method{} predicts electrophysiology from the transcriptome most accurately on all three cohorts (Table~\ref{tab:forward}). It has the highest correlation, the highest AUROC and the lowest error in mouse visual cortex, in mouse motor cortex and in human cortex. Against JAMIE, an existing cross-modal Patch-seq imputation method, it is more accurate on 36 of 39 features in visual cortex and on 27 of 29 in motor cortex. On the human cohort, which has 612 cells, JAMIE is the closest baseline in correlation ($0.406$ against $0.414$), and \method{} has the higher AUROC on 50 of 75 features (per-feature comparison with every baseline in Table~\ref{tab:app-wins}). %

Single-cell foundation models pretrained on tens of millions of cells are less accurate on these cohorts than the language-based representation. MLPs on frozen Geneformer, UCE and Nicheformer embeddings reach a correlation of at most $0.492$ in visual cortex and $0.503$ in motor cortex, against $0.522$ and $0.520$ for \method{}, and Geneformer falls to $0.146$ in motor cortex. Training the last transformer block of Nicheformer with its head changes its correlation by less than $0.01$ on every cohort; with about $10^3$ training cells it overfits (Appendix~\ref{app:hyper}). %

The closest baseline on the mouse cohorts is the Plain GenePT MLP ($0.505$ in visual and $0.502$ in motor cortex), which uses the same gene embeddings without the adapter and the alignment. \method{} is more accurate than this baseline on 36 of 39, 22 of 29 and 48 of 75 features in the three cohorts. The gain is attributable to the representation: evaluated with the Plain GenePT baseline's own head, the \method{} embedding reaches $0.522$ and $0.531$ (Sec.~\ref{sec:results-ablation}). %

\subsection{Shared space: alignment and reverse readouts}
\label{sec:results-reverse}

\begin{figure}[!b]\centering\includegraphics[width=\linewidth]{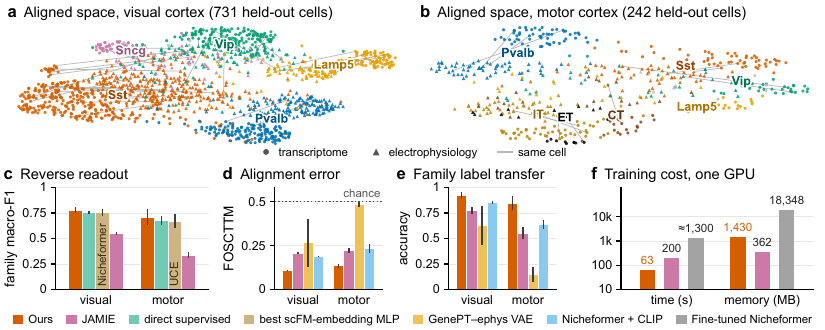}\caption{Shared space. (a, b) UMAP of the aligned space (held-out cells of one fold), coloured by transcriptomic family. (c) Reverse readout of transcriptomic family from electrophysiology (macro-F1). (d) Held-out FOSCTTM (lower is better) and (e) family label-transfer accuracy for the methods with a joint space, JAMIE re-run on our folds. (f) Training time and peak GPU memory for one fold (Appendix~\ref{app:integration}). Error bars, s.d.\ over five folds.}\label{fig:reverse}\end{figure}

\method{} aligns the two modalities more closely than every other method with a joint space (Fig.~\ref{fig:reverse}d,e). On held-out cells, FOSCTTM is $0.107$ in visual cortex and $0.135$ in motor cortex, against $0.208$ and $0.222$ for JAMIE and $0.187$ and $0.235$ for Nicheformer+CLIP. Family label-transfer accuracy is $0.926$ and $0.848$, against $0.777$ and $0.551$ for JAMIE (Table~\ref{tab:integration}). Alignment trains in 63\,s on one GPU, against 200\,s for JAMIE and about 1{,}300\,s for fine-tuning Nicheformer (Fig.~\ref{fig:reverse}f). %

The aligned space also supports the reverse direction, in which transcriptomic identity is predicted from electrophysiology alone (Fig.~\ref{fig:reverse}c; Table~\ref{tab:reverse}). One multitask head on the aligned representation predicts the transcriptomic family (macro-F1 $0.779$ in visual and $0.710$ in motor cortex), the fine type, the cortical layer, the expression of 27 marker genes ($r=0.539$ and $0.572$) and 9 marker programs. Compared with classifiers and regressors trained on the electrophysiological features for each task, it is more accurate on 4 of 5 endpoints in visual cortex and on all 5 in motor cortex. It exceeds JAMIE's reverse imputation on every endpoint (JAMIE: family macro-F1 $0.549$ and $0.336$). %

\subsection{Transfer across cortical areas, laboratories and species}
\label{sec:results-transfer}

\begin{figure}[t]
\centering
\includegraphics[width=\linewidth]{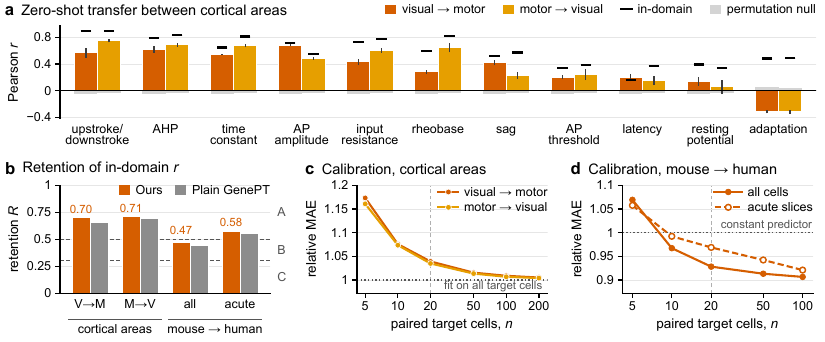}
\caption{Zero-shot transfer without target training.
(a) Per-concept Pearson correlations for visual$\leftrightarrow$motor transfer, evaluated using the cross-area matched electrophysiology features and permutation-based chance levels.
(b) Correlation retention, defined relative to the corresponding in-domain performance, for cross-area and mouse-to-human transfer.
(c,d) Few-shot calibration as a function of the number of paired target cells for cross-area and mouse-to-human transfer.}
\label{fig:transfer}
\end{figure}

\paragraph{Cortical areas.}
A model trained on one mouse cortical area predicts electrophysiology in the other without retraining (Fig.~\ref{fig:transfer}a,b). It retains $70\%$ of the in-domain correlation from visual to motor cortex and $71\%$ in the opposite direction, and all 11 concepts exceed the permutation null in both directions. Spike shape and passive membrane properties transfer best and resting potential worst; the adaptation index correlates negatively because the two pipelines define it with opposite signs. The mean concept correlation is $0.344$ and $0.384$ for \method{}, $0.317$ and $0.371$ for the Plain GenePT MLP and $0.326$ and $0.385$ for the Nicheformer MLP, whereas Geneformer does not transfer ($0.05$ and $0.04$; Appendix~\ref{app:transfer-tables}). Predictions stay on the source pipeline's scale, and affine calibration on 20 paired target cells brings the held-out error within $4\%$ of calibration on all target cells (Fig.~\ref{fig:transfer}c). %

\paragraph{Laboratories.}
Because a recording enters the model as text, a trained model can be applied to recordings described by a different feature set. The two mouse cohorts come from different laboratories and share no feature names (39 IPFX features; 29 features of \citealp{scala2021phenotypic}). We applied the visual-cortex model to the motor-cortex recordings, verbalized in their own 29 features, without retraining and without a feature correspondence, and measured label transfer between the recording and transcriptome embeddings of the same cells. Family label-transfer accuracy is $0.51$ against a permutation null of $0.19$ ($0.78$ against $0.30$ on the interneuron families, the classes present in the training cohort), and $0.65$ against $0.31$ in the opposite direction (Fig.~\ref{fig:lang}c). This setting is infeasible for an encoder with one input unit per feature, which is tied to the feature set of its training cohort and requires a manually constructed feature correspondence or retraining. %

\paragraph{Species.}
Transfer also holds between mouse and human cortex, in both directions (Fig.~\ref{fig:lang}d). Without any human training, the mouse visual-cortex model exceeds the permutation null on 18 of 25 concepts in human interneurons and retains $47\%$ of the in-domain correlation ($58\%$ on the 247 acute-slice cells). Conversely, a model trained on the 612 human neurons and applied to the 3{,}654 mouse visual-cortex neurons exceeds the null on 23 of 24 concepts and retains $63\%$ of the in-domain correlation (Table~\ref{tab:app-h2m}). Transfer is stronger toward mouse although the human training cohort is six times smaller, which suggests that mouse-to-human transfer is limited by the variability of the human recordings more than by the species difference. Permuting the gene identities of the input reduces the mean $r$ to $0.00$ and $-0.02$ in the two directions, and within human subclasses the mean $r$ is $0.17$ ($0.30$ over all cells), so the predictions use gene-level information beyond subclass means. %

\paragraph{Few-shot adaptation.}
A small number of paired human cells suffices to adapt the mouse-trained model. An affine calibration on 20 paired human cells gives a held-out error within $2.5\%$ of a calibration on 100 cells (Fig.~\ref{fig:transfer}d). With 25 labelled human cells, a ridge head on the frozen mouse-trained representation reaches $r=0.270$, against $0.197$ for Plain GenePT embeddings and $0.195$ for Geneformer embeddings; with 50 cells it reaches $0.297$ against $0.264$ and $0.256$ (Table~\ref{tab:app-lowresource}). %

\subsection{Ablations}
\label{sec:results-ablation}

\begin{table}[t]\centering
\caption{Ablations on held-out cells (five-fold means). Every row, including the full model, retrains the alignment stage with all other settings fixed (Appendix~\ref{app:ablations}). (a) Text or numerical encoder for each modality under one shared readout head; lower rows, readout head of the Plain GenePT MLP. (b) Components of the alignment stage, per-feature heads of Table~\ref{tab:forward}.}
\label{tab:ablation}
\small
\setlength{\tabcolsep}{3pt}
\resizebox{\textwidth}{!}{%
\begin{tabular}{ll ccc ccc}
\toprule
\multicolumn{2}{l}{(a) Encoder of each modality} & \multicolumn{3}{c}{Mouse visual} & \multicolumn{3}{c}{Mouse motor} \tabularnewline
\cmidrule(lr){3-5}\cmidrule(lr){6-8}
Expression & Electrophysiology & $r\uparrow$ & AUROC$\uparrow$ & MAE$\downarrow$ & $r\uparrow$ & AUROC$\uparrow$ & MAE$\downarrow$ \tabularnewline
\midrule
text (GenePT + adapter) & text (verbalized record): \method{} & \textbf{0.521} & \textbf{0.777} & \textbf{9.09} & \textbf{0.532} & \textbf{0.807} & \textbf{7.90} \tabularnewline
text (GenePT, no adapter) & numerical (MLP on quintile bins) & 0.507 & 0.767 & 9.24 & 0.495 & 0.783 & 8.17 \tabularnewline
numerical (linear) & text (verbalized record) & 0.501 & 0.771 & 9.18 & 0.517 & 0.786 & 8.12 \tabularnewline
numerical (MLP) & text (verbalized record) & 0.472 & 0.754 & 9.62 & 0.503 & 0.785 & 8.53 \tabularnewline
numerical (MLP) & numerical (MLP on quintile bins) & 0.419 & 0.717 & 10.61 & 0.495 & 0.782 & 8.66 \tabularnewline
\midrule
\multicolumn{2}{l}{Plain GenePT embedding, its own readout head (Table~\ref{tab:forward})} & 0.505 & 0.768 & 9.22 & 0.502 & 0.790 & 8.02 \tabularnewline
\multicolumn{2}{l}{\method{} embedding, readout head of the Plain GenePT MLP} & \textbf{0.522} & \textbf{0.778} & \textbf{9.05} & \textbf{0.531} & \textbf{0.807} & \textbf{7.87} \tabularnewline
\bottomrule
\end{tabular}
}
\par\medskip
\resizebox{\textwidth}{!}{%
\begin{tabular}{l cccc cccc}
\toprule
 & \multicolumn{4}{c}{Mouse visual} & \multicolumn{4}{c}{Mouse motor} \tabularnewline
\cmidrule(lr){2-5}\cmidrule(lr){6-9}
(b) Components of the alignment stage & $r\uparrow$ & MAE$\downarrow$ & FOSCTTM$\downarrow$ & LTA$\uparrow$ & $r\uparrow$ & MAE$\downarrow$ & FOSCTTM$\downarrow$ & LTA$\uparrow$ \tabularnewline
\midrule
\method{} (all four components) & \textbf{0.521} & \textbf{8.98} & \textbf{0.109} & \textbf{0.927} & \textbf{0.521} & \textbf{7.65} & \textbf{0.139} & \textbf{0.822} \tabularnewline
without the context adapter & 0.516 & 9.07 & 0.110 & 0.921 & 0.501 & 7.90 & 0.159 & 0.745 \tabularnewline
adapter with label-free statistics only & 0.519 & 9.07 & 0.113 & 0.912 & 0.505 & 7.92 & 0.151 & 0.780 \tabularnewline
without all four components (contrastive loss only) & 0.519 & 9.04 & 0.118 & 0.915 & 0.506 & 7.92 & 0.170 & 0.800 \tabularnewline
\bottomrule
\end{tabular}
}
\end{table}

\begin{figure}[t]\centering\includegraphics[width=\linewidth]{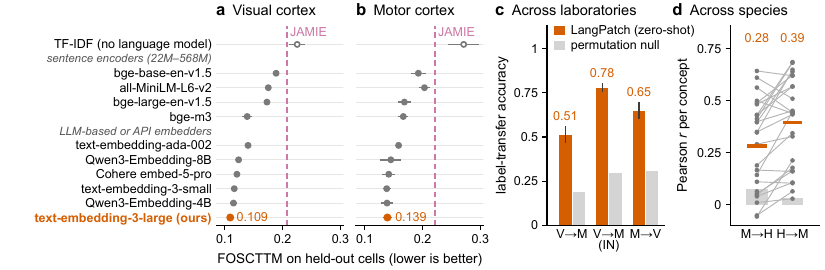}\caption{Text embedders and zero-shot generalization. (a, b) Held-out alignment error of \method{} retrained with each of eleven text embedders, grouped by type; open circle, bag of words; dashed line, JAMIE. (c) Family label transfer between recordings described by another laboratory's features and transcriptomes, with its permutation null; V$\to$M, visual-cortex model on motor-cortex recordings; M$\to$V, the reverse; IN, interneurons only. (d) Zero-shot transfer between mouse (M) and human (H): Pearson $r$ per concept, lines join the same concept; orange, means; grey bars, permutation null. Error bars, s.d.\ over five folds (a, b) or fold models (c).}\label{fig:lang}\end{figure}

\paragraph{Language interface.}
Every variant in which a numerical encoder trained on the cohort replaces a text embedding predicts less accurately than \method{} (Table~\ref{tab:ablation}a; same training objective and readout head in every row). With numerical encoders for both modalities, which is contrastive learning without the language model, the correlation falls from $0.521$ to $0.419$ in visual cortex and from $0.532$ to $0.495$ in motor cortex.

\paragraph{Text embedder.}
Alignment improves with the quality of the text embedder (Fig.~\ref{fig:lang}a,b; Table~\ref{tab:app-embedders}). With the gene descriptions and the records re-embedded by ten other embedders and all other settings fixed, FOSCTTM in visual cortex falls from $0.225$ for a bag-of-words representation without a language model to $0.109$ for the default embedder, and from $0.271$ to $0.139$ in motor cortex; every language-model embedder aligns better than JAMIE in both cohorts. The cross-laboratory test shows the same trend: the five embedders with the lowest FOSCTTM reach a mean label-transfer accuracy of $0.49$ and $0.55$ in the two directions, against $0.35$ and $0.29$ for the other six. \method{} therefore benefits directly from progress in general-purpose text embedders. %

\paragraph{Components.}
Among the four components added to the contrastive objective, the context adapter contributes most (Table~\ref{tab:ablation}b). Removing it lowers the prediction correlation on 31 of 39 features in visual cortex and on 23 of 29 in motor cortex ($p<0.001$ for both) and raises FOSCTTM in motor cortex from $0.139$ to $0.159$; on the human cohort (612 cells), FOSCTTM rises from $0.283$ to $0.312$. An adapter restricted to label-free gene statistics retains most of the margin over JAMIE (FOSCTTM $0.113$ and $0.151$, against $0.208$ and $0.222$). Removing all four components, which leaves contrastive learning on the frozen embeddings, raises FOSCTTM in every fold of both cohorts. Appendix~\ref{app:ablations} reports the other single removals and the sensitivity to the four hyperparameters: over eight settings, FOSCTTM stays within $0.102$--$0.117$ in visual cortex and $0.131$--$0.154$ in motor cortex. %

\paragraph{Full method and its ingredients.}
\method{} outperforms each of its ingredients used alone: GenePT embeddings with an MLP head under a matched readout head (Table~\ref{tab:ablation}a, lower rows), contrastive learning with numerical encoders (Table~\ref{tab:ablation}a, fifth row), and GenePT embeddings, verbalized records and the contrastive loss without our components (Table~\ref{tab:ablation}b, last row). %

\section{Discussion}
\label{sec:discussion}
\paragraph{What the shared space encodes.}
In the aligned space, held-out cells of both modalities cluster by transcriptomic family (Fig.~\ref{fig:reverse}a,b). Because the cell embedding is linear in expression, each forward head can be attributed to individual genes, and we compare the resulting gene ranking with those of the Plain GenePT MLP and JAMIE under the same gradient$\times$input protocol. Of the cell-type markers defined on the training fold, 64\% lie among the top 200 genes of \method{} in visual cortex and 51\% in motor cortex (random ranking: 21\% and 20\%; Plain GenePT MLP: 64\% and 37\%; JAMIE: 5\% and 12\%). Ion-channel and excitability genes are enriched 2.8-fold and 2.9-fold among its top 50 genes (95th percentile of random gene sets: 2.6-fold), and removing these genes from the transcriptome input lowers the label-transfer accuracy of the joint space by 0.02 and 0.05, against at most 0.004 for random genes. %
The Plain GenePT MLP loses more accuracy than \method{} when each model's top-ranked genes are deleted, and its top genes are more enriched for synaptic-transmission genes. The gene$\times$feature maps and the full benchmark are in Appendix~\ref{app:attribution}; attribution describes what the predictors rely on and is hypothesis-generating.

\paragraph{Limitations.}
Our evidence is limited to two mouse cortical areas and one primary human cohort, and only the motor cortex cohort contains excitatory neurons. Zero-shot transfer primarily preserves relative variation; a small number of paired target cells may still be needed for calibration, while recording-sensitive features like threshold voltage transfer poorly even after calibration. The default text embedder, OpenAI's text-embedding-3-large, is available only through an API; the open-weight Qwen3-Embedding-4B gives a similar alignment (Table~\ref{tab:app-embedders}).

\paragraph{Conclusion.}
A frozen general-purpose language model can serve as an interface between transcriptome and electrophysiology. A lightweight contrastive framework that adapts a pretrained gene-language prior and aligns it with verbalized electrophysiology, trained on only about $10^3$ paired cells, gives the most accurate transcriptome-to-electrophysiology prediction among the evaluated foundation-model and representation-learning methods across three cohorts, the best cross-modal embedding alignment on both mouse cohorts, reverse electrophysiology-to-transcriptome predictions that exceed the baselines on most endpoints, and gene attributions that recover cell-type markers. In principle, the same frozen text embedder and contrastive objective apply to any modality that can be verbalized as named quantities, such as morphological features or chromatin accessibility in single-cell multiome data~\citep{cohenkalafut2023jamie}. One representation thus serves atlas cells that will never be patched and recordings that will never be sequenced: a building block for multimodal foundation models in neuroscience.

%% file: 4_statements.tex
\makeatletter
\write\@auxout{\string\gdef\string\refstart{\thepage}}
\makeatother

\subsection*{AI use statement}
Generative AI tools were used in this work. AI coding assistants based on large language models were used, under the authors' direction and to the authors' specifications, to write and refactor training, evaluation, and analysis scripts. The authors designed the method, chose the datasets, baselines and metrics, wrote the evaluation criteria, and reviewed all AI-assisted code and text. Every number reported in the paper is regenerated by the released scripts from the released result files (CSV), so no reported value originates from the free-text output of an AI tool; no dataset, measurement or result was produced by a generative model, and all inputs are the public datasets listed in the reproducibility statement. We take full responsibility for the final content of this work, including text, claims and artifacts produced with the aid of AI.

\subsection*{Ethics statement}
All data used in this work are public and de-identified: the mouse Patch-seq cohorts of \citet{gouwens2020integrated} and \citet{scala2021phenotypic}, and the human cortical Patch-seq cohorts of \citet{lee2023signature} and \citet{chartrand2023morphoelectric}. The authors performed no experiments on animals or human subjects; the human tissue data were collected and consented under the original studies' protocols, and we use only the processed, de-identified tables released by those studies. The method produces predicted physiological properties from transcriptomes; such predictions are model outputs, not measurements. We see no direct path from this work to harmful applications; the main risk is over-interpretation of imputed electrophysiology, which we address by reporting permutation nulls, negative controls and failure cases alongside the positive results.

\subsection*{Reproducibility statement}
Code and the result files (CSV) underlying the tables and figures are available at \url{https://github.com/ai4biomedicine/LangPatch}. All datasets are public: the mouse visual cortex cohort of \citet{gouwens2020integrated} (Allen Institute; raw electrophysiology in DANDI dandiset 000020), the mouse motor cortex cohort of \citet{scala2021phenotypic} (berenslab mini-atlas), and the human cohorts of \citet{lee2023signature} and \citet{chartrand2023morphoelectric} (processed tables from the Allen Institute GitHub releases). Both mouse cohorts are the scMNC preprocessing used by JAMIE~\citep{cohenkalafut2023jamie}, so cell counts and folds are directly comparable with that work. The gene embeddings are the released GenePT v2 vectors~\citep{chen2024genept} (OpenAI \texttt{text-embedding-3-large}, 3072 dimensions~\citep{openai2024embeddings}, applied by the GenePT authors to NCBI gene summaries and UniProt protein summaries; Zenodo record 10833191); the verbalized electrophysiology records were embedded once by us with the same model.  Appendix~\ref{app:hyper} lists every hyperparameter, parameter count and model-selection rule; Appendix~\ref{app:text} the verbalization template; Appendix~\ref{app:crosswalk} the crosswalk, permutation-null, retention and calibration protocols; Appendix~\ref{app:integration} the integration-metric and timing protocols; Appendix~\ref{app:human-indomain} the human fold design and nested validation; Appendix~\ref{app:attribution} the attribution and visualization protocols; and Appendix~\ref{app:ablations} the ablation, cross-laboratory and human-to-mouse protocols.

%% file: appendix.tex
\section{Ablations, cross-laboratory alignment and human-to-mouse transfer}
\label{app:ablations}

\paragraph{Protocol.}
Every model in this appendix is trained with the stage-1 procedure of Appendix~\ref{app:hyper} (same objective, schedule, hyperparameters and checkpoint rule) on the same five folds, with one run per configuration and no tuning; only the stated element changes. FOSCTTM and family label-transfer accuracy (LTA) are computed on the held-out test cells as in Appendix~\ref{app:integration}, and forward prediction uses the per-feature heads of Table~\ref{tab:forward} (512--128 units) unless stated otherwise. The default configuration was retrained together with the variants; its FOSCTTM of 0.109 and 0.139 differs slightly from the 0.107 and 0.135 of Table~\ref{tab:integration}.

\begin{table}[h]\centering
\caption{Alignment stage: component ablation and hyperparameter sensitivity (five-fold means on held-out cells). (a) Each row removes one of the four components that \method{} adds to the contrastive loss; ``without all four'' is contrastive learning on the frozen embeddings. (b) Each of the four hyperparameters is set to a lower and a higher value with the other three at their defaults.}
\label{tab:app-components}
\small
\setlength{\tabcolsep}{3pt}
\resizebox{\textwidth}{!}{%
\begin{tabular}{l cccc cccc}
\toprule
 & \multicolumn{4}{c}{Mouse visual} & \multicolumn{4}{c}{Mouse motor} \tabularnewline
\cmidrule(lr){2-5}\cmidrule(lr){6-9}
(a) Model & $r\uparrow$ & MAE$\downarrow$ & FOSCTTM$\downarrow$ & LTA$\uparrow$ & $r\uparrow$ & MAE$\downarrow$ & FOSCTTM$\downarrow$ & LTA$\uparrow$ \tabularnewline
\midrule
\method{} (all four components) & 0.521 & 8.98 & 0.109 & 0.927 & 0.521 & 7.65 & 0.139 & 0.822 \tabularnewline
without (1) the context adapter & 0.516 & 9.07 & 0.110 & 0.921 & 0.501 & 7.90 & 0.159 & 0.745 \tabularnewline
\quad adapter with label-free statistics only & 0.519 & 9.07 & 0.113 & 0.912 & 0.505 & 7.92 & 0.151 & 0.780 \tabularnewline
without (2) the gene-cosine penalty & 0.519 & 9.04 & 0.112 & 0.921 & 0.522 & 7.68 & 0.135 & 0.873 \tabularnewline
without (3) the cell-preserve penalty & 0.521 & 9.01 & 0.108 & 0.923 & 0.520 & 7.61 & 0.143 & 0.809 \tabularnewline
without (4) the auxiliary regression, transcriptome side & 0.521 & 9.03 & 0.110 & 0.926 & 0.518 & 7.73 & 0.136 & 0.833 \tabularnewline
without (4) the auxiliary regression, recording side and its warm-up & 0.522 & 9.01 & 0.116 & 0.929 & 0.526 & 7.68 & 0.143 & 0.839 \tabularnewline
without (4) the auxiliary regression, both sides & 0.524 & 8.99 & 0.114 & 0.927 & 0.529 & 7.65 & 0.140 & 0.850 \tabularnewline
without all four & 0.519 & 9.04 & 0.118 & 0.915 & 0.506 & 7.92 & 0.170 & 0.800 \tabularnewline
\midrule
JAMIE (Tables~\ref{tab:forward} and~\ref{tab:integration}) & 0.453 & 10.05 & 0.208 & 0.777 & 0.432 & 8.91 & 0.222 & 0.551 \tabularnewline
\bottomrule
\end{tabular}
}
\par\medskip
\begin{tabular}{l c cc cc}
\toprule
 & & \multicolumn{2}{c}{FOSCTTM$\downarrow$} & \multicolumn{2}{c}{LTA$\uparrow$} \tabularnewline
\cmidrule(lr){3-4}\cmidrule(lr){5-6}
(b) Hyperparameter & Value & visual & motor & visual & motor \tabularnewline
\midrule
Default setting & & 0.109 & 0.139 & 0.927 & 0.822 \tabularnewline
\midrule
Gene-cosine weight $\lambda_g$ (default 0.5) & 0.25 & 0.110 & 0.134 & 0.926 & 0.841 \tabularnewline
 & 1.0 & 0.111 & 0.139 & 0.925 & 0.817 \tabularnewline
Cell-preserve weight $\lambda_c$ (default 0.1) & 0.05 & 0.111 & 0.136 & 0.934 & 0.816 \tabularnewline
 & 0.2 & 0.108 & 0.137 & 0.931 & 0.825 \tabularnewline
Temperature $\tau$ (default 0.07) & 0.03 & 0.117 & 0.154 & 0.934 & 0.790 \tabularnewline
 & 0.15 & 0.102 & 0.131 & 0.924 & 0.878 \tabularnewline
Gate scale $s$ (default 0.1) & 0.05 & 0.111 & 0.140 & 0.922 & 0.782 \tabularnewline
 & 0.2 & 0.110 & 0.133 & 0.924 & 0.871 \tabularnewline
\bottomrule
\end{tabular}
\end{table}

\paragraph{Components and hyperparameters.}
Table~\ref{tab:app-components}a removes each of the four components that \method{} adds to the contrastive loss: (1) the context adapter, (2) the gene-cosine penalty, (3) the cell-preserve penalty and (4) the auxiliary regression. The context adapter contributes most. Without it, the prediction correlation is lower on 31 of 39 features in visual cortex ($p=1\times10^{-4}$, paired Wilcoxon test over features) and on 23 of 29 in motor cortex ($p=2\times10^{-5}$). In motor cortex FOSCTTM rises from 0.139 to 0.159 and LTA falls from 0.822 to 0.745, and on the human cohort FOSCTTM rises from 0.283 to 0.312 and LTA falls from 0.575 to 0.521. An adapter that sees only the four label-free gene statistics (mean, standard deviation, detection rate and coefficient of variation) gives a FOSCTTM of 0.113 and 0.151. The gene-cosine penalty improves prediction in visual cortex (27 of 39 features are lower without it, $p=0.005$). The auxiliary regression on the recording side lowers FOSCTTM in visual cortex in every fold (0.109 against 0.116 without it), and the cell-preserve penalty lowers it in motor cortex (0.139 against 0.143). Removing all four components raises FOSCTTM in every fold of both cohorts and lowers the prediction correlation on 25 of 39 features in visual cortex ($p=0.049$) and on 22 of 29 in motor cortex ($p=5\times10^{-4}$). The alignment stage has four hyperparameters: the weights of the two cosine penalties, the contrastive temperature and the gate scale of the adapter. Across the eight settings of Table~\ref{tab:app-components}b, FOSCTTM stays between 0.102 and 0.117 in visual cortex and between 0.131 and 0.154 in motor cortex, compared with 0.208 and 0.222 for JAMIE.

\paragraph{Text or numerical encoders.}
In the upper rows of Table~\ref{tab:ablation}a the text embedding of a modality is replaced by a numerical encoder trained from scratch on the cohort with the objective, schedule and checkpoint rule of \method{}. The numerical expression encoder reads the log-normalized expression vector over the gene panel, the same input over which \method{} pools gene embeddings, through an MLP (layer normalization, linear, GELU, linear, layer normalization) or through a linear map of the form of $P_{\mathrm{cell}}$. The numerical recording encoder reads the one-hot quintile labels of the features, the binned information that the verbalized record also contains, through an MLP of the same form. These rows are evaluated with one multi-output MLP head at the default setting of the single-cell foundation-model baselines (Appendix~\ref{app:hyper}). In the row that combines GenePT with an MLP on quintile bins, the transcriptome side is the plain GenePT cell embedding without the adapter; trained with the contrastive loss alone, the same model reaches $r=0.516$, AUROC 0.774 and MAE 9.13 in visual cortex and $r=0.514$, AUROC 0.794 and MAE 8.06 in motor cortex.

\paragraph{Matched readout heads.}
In Table~\ref{tab:forward}, \method{} uses per-feature heads whose width is selected on the validation fold, and the embedding baselines use one multi-output head at default settings. The lower rows of Table~\ref{tab:ablation}a evaluate the \method{} embedding with the head of the Plain GenePT MLP, trained with that baseline's settings and without any selection: the margin in $r$ over Plain GenePT is 0.017 in visual cortex and 0.029 in motor cortex (0.017 and 0.018 in Table~\ref{tab:forward}). With the head of the single-cell foundation-model baselines (first row of Table~\ref{tab:ablation}a), the \method{} embedding reaches $r=0.521$ and $0.532$, against $0.486$ and $0.503$ for Nicheformer and $0.492$ and $0.146$ for Geneformer under the same head (Table~\ref{tab:forward}).

\begin{table}[h]\centering
\caption{Text-embedder ablation. Gene descriptions and records are re-embedded with each embedder and \method{} is retrained with all other settings fixed (five-fold means on held-out cells). Cross-laboratory LTA is the zero-shot test of Fig.~\ref{fig:lang}c (visual-cortex model on motor-cortex recordings, and the reverse). Rows are grouped by embedder type as in Fig.~\ref{fig:lang} and ordered by FOSCTTM in visual cortex within each group.}
\label{tab:app-embedders}
\small
\setlength{\tabcolsep}{3pt}
\resizebox{\textwidth}{!}{%
\begin{tabular}{l cc cc cc cc}
\toprule
 & \multicolumn{2}{c}{FOSCTTM$\downarrow$} & \multicolumn{2}{c}{LTA$\uparrow$} & \multicolumn{2}{c}{Forward MAE$\downarrow$} & \multicolumn{2}{c}{Cross-laboratory LTA$\uparrow$} \tabularnewline
\cmidrule(lr){2-3}\cmidrule(lr){4-5}\cmidrule(lr){6-7}\cmidrule(lr){8-9}
Text embedder (used for both modalities) & visual & motor & visual & motor & visual & motor & vis.$\to$mot. & mot.$\to$vis. \tabularnewline
\midrule
TF-IDF + SVD (bag of words, no language model) & 0.225 & 0.271 & 0.835 & 0.684 & 9.05 & 7.77 & 0.31 & 0.48 \tabularnewline
\midrule
\multicolumn{9}{l}{\textit{Sentence encoders}} \tabularnewline
bge-base-en-v1.5 (open, 110M) & 0.189 & 0.193 & 0.899 & 0.772 & 9.04 & 7.70 & 0.41 & 0.16 \tabularnewline
all-MiniLM-L6-v2 (open, 22M) & 0.175 & 0.203 & 0.902 & 0.699 & 9.09 & 7.78 & 0.39 & 0.35 \tabularnewline
bge-large-en-v1.5 (open, 335M) & 0.173 & 0.169 & 0.923 & 0.770 & 9.02 & 7.73 & 0.36 & 0.24 \tabularnewline
bge-m3 (open, 568M) & 0.139 & 0.167 & 0.913 & 0.780 & 9.02 & 7.61 & 0.37 & 0.36 \tabularnewline
\midrule
\multicolumn{9}{l}{\textit{LLM-based or API embedders}} \tabularnewline
text-embedding-ada-002 (API) & 0.140 & 0.159 & 0.918 & 0.819 & 9.06 & 7.69 & 0.27 & 0.14 \tabularnewline
Qwen3-Embedding-8B (open) & 0.124 & 0.145 & 0.936 & 0.822 & 9.06 & 7.64 & 0.46 & 0.58 \tabularnewline
Cohere embed-5-pro (API) & 0.121 & 0.141 & 0.924 & 0.825 & 9.03 & 7.69 & 0.57 & 0.53 \tabularnewline
text-embedding-3-small (API) & 0.116 & 0.138 & 0.927 & 0.834 & 9.02 & 7.72 & 0.40 & 0.60 \tabularnewline
Qwen3-Embedding-4B (open) & 0.115 & 0.139 & 0.930 & 0.809 & 9.01 & 7.70 & 0.49 & 0.39 \tabularnewline
text-embedding-3-large (API; \method{}) & 0.109 & 0.139 & 0.927 & 0.822 & 8.98 & 7.65 & 0.54 & 0.65 \tabularnewline
\bottomrule
\end{tabular}
}
\end{table}

\paragraph{Text embedders.}
The gene summaries and the verbalized records of both mouse cohorts are re-embedded with ten other embedders, and both stages of \method{} are retrained with all other settings fixed (Table~\ref{tab:app-embedders}; Fig.~\ref{fig:lang}a,b). The embedders are a bag-of-words representation without a language model (TF-IDF vectors of the texts reduced to 1{,}024 dimensions by truncated SVD), four open-weight sentence encoders (all-MiniLM-L6-v2, \citealp{reimers2019sentence,wang2020minilm}; bge-base-en-v1.5 and bge-large-en-v1.5, \citealp{xiao2024cpack}; bge-m3, \citealp{chen2024bgem3}), two open-weight embedders built on large language models (Qwen3-Embedding-4B and 8B; \citealp{zhang2025qwen3}) and four models served through an API (text-embedding-ada-002, text-embedding-3-small and text-embedding-3-large, \citealp{openai2024embeddings}; Cohere embed-5-pro). Texts longer than the context of an embedder are split into chunks whose embeddings are averaged. FOSCTTM falls from 0.225 and 0.271 with the bag-of-words representation to 0.109 and 0.139 with the default embedder. Each of the ten language-model embedders gives a lower FOSCTTM than JAMIE in both cohorts (at most 0.189 and 0.203, against 0.208 and 0.222), whereas the bag-of-words representation gives a higher one. In the cross-laboratory test, the five embedders with the lowest FOSCTTM reach a mean label-transfer accuracy of 0.49 (visual-cortex model on motor-cortex recordings) and 0.55 (the reverse), against 0.35 and 0.29 for the other six; the permutation null of this test is 0.19 and 0.31 for the default embedder.

\paragraph{Cross-laboratory alignment.}
The two mouse cohorts were recorded in different laboratories and their feature sets share no names (39 IPFX features; the 29 features of \citealp{scala2021phenotypic}). The five fold models of the visual-cortex cohort are applied unchanged to the 1{,}208 cells of the motor-cortex cohort. On the recording side, each motor-cortex record is verbalized with its own 29 feature names and units (Appendix~\ref{app:text}), embedded by the text embedder and mapped by the visual-cortex model's $P_{\mathrm{ephys}}$. On the transcriptome side, the expression of the cell over the visual-cortex gene panel is passed through the visual-cortex model's adapted gene matrix and $P_{\mathrm{cell}}$. No parameter is trained, and no feature of one cohort is matched to a feature of the other. Fig.~\ref{fig:lang}c reports family label transfer between the two embeddings of the same cells ($k=5$, as in Appendix~\ref{app:integration}), with the mean of 1{,}000 permutations of the cell pairing as the null. The accuracy is 0.51 on the 1{,}208 cells (null 0.19; share of the largest family 0.24). The motor-cortex cohort contains excitatory families that are absent from the visual-cortex cohort, and on its 817 interneurons the accuracy is 0.78 (null 0.30; largest family 0.35). For the motor-cortex model applied to the 3{,}654 visual-cortex cells it is 0.65 (null 0.31; largest family 0.42). FOSCTTM is 0.41, 0.39 and 0.41 in the three tests, against 0.50 at chance. Every test exceeds the 95th percentile of the permutation null for all five fold models (permutation $p=0.001$).

\begin{table}[h]\centering
\caption{Human-to-mouse zero-shot transfer per concept: Pearson $r$ between the prediction of the human-trained model and the measured value on the 3{,}654 mouse visual-cortex cells (H$\to$M), the mouse-to-human value of the same concept (M$\to$H), and the in-domain held-out $r$ of the mouse visual-cortex model (in-dom.). LS, long square; SS, short square. $^{\circ}$Not above the 95th percentile of the permutation null.}
\label{tab:app-h2m}
\small
\setlength{\tabcolsep}{3pt}
\resizebox{\textwidth}{!}{%
\begin{tabular}{l ccc l ccc}
\toprule
Concept & H$\to$M & M$\to$H & in-dom. & Concept & H$\to$M & M$\to$H & in-dom. \tabularnewline
\midrule
Fast AHP (LS) & 0.68 & 0.43 & 0.83 & Potential at the sag step & 0.44 & 0.50 & 0.55 \tabularnewline
Trough (LS) & 0.68 & 0.41 & 0.79 & AP amplitude (LS) & 0.43 & 0.35 & 0.55 \tabularnewline
Rheobase & 0.65 & 0.35 & 0.83 & Mean inter-spike interval & 0.34 & 0.19 & 0.55 \tabularnewline
Input resistance & 0.64 & 0.40 & 0.78 & Upstroke:downstroke ratio (SS) & 0.33 & 0.29 & 0.75 \tabularnewline
Fast AHP (ramp) & 0.62 & 0.49 & 0.76 & Fast AHP (SS) & 0.21 & $-$0.06 & 0.62 \tabularnewline
Membrane time constant & 0.62 & 0.64 & 0.81 & AP threshold (LS) & 0.18 & 0.05 & 0.39 \tabularnewline
Threshold current (ramp) & 0.57 & 0.43 & 0.76 & Resting potential & 0.16 & 0.05 & 0.34 \tabularnewline
Trough (ramp) & 0.57 & 0.48 & 0.69 & AP threshold (ramp) & 0.16 & 0.17 & 0.30 \tabularnewline
Upstroke:downstroke ratio (LS) & 0.51 & 0.61 & 0.90 & AP threshold (SS) & 0.11 & 0.04 & 0.34 \tabularnewline
AP amplitude (ramp) & 0.45 & 0.40 & 0.53 & Sag ratio & 0.10 & 0.11 & 0.57 \tabularnewline
f--I slope & 0.45 & 0.14 & 0.76 & AP amplitude (SS) & 0.06 & $-$0.05 & 0.37 \tabularnewline
Upstroke:downstroke ratio (ramp) & 0.44 & 0.56 & 0.86 & Latency & 0.03$^{\circ}$ & 0.01 & 0.37 \tabularnewline
\bottomrule
\end{tabular}
}
\end{table}

\paragraph{Human-to-mouse transfer.}
\method{} is trained on the 612 cells of the human cohort only (the alignment stage and the validation-selected heads of Table~\ref{tab:forward}; five fold models) and applied without adaptation to the 3{,}654 cells of the mouse visual-cortex cohort, whose expression is given over the same 1{,}293-gene panel. Scoring follows Appendix~\ref{app:crosswalk} with the roles of the two species exchanged: the prediction of the human feature of each concept is correlated with the measured mouse feature, the null is the 95th percentile of 1{,}000 cell permutations (0.031 to 0.035), and retention is the mean $|r|$ divided by the in-domain held-out mean $|r|$ of the mouse visual-cortex model over the same concepts (0.625). The adaptation index is not among the 75 human features, which leaves 24 of the 25 concepts. The mean $r$ is 0.39, 23 of the 24 concepts exceed the null, and retention is 63\% (Table~\ref{tab:app-h2m}; Fig.~\ref{fig:lang}d); the correlation is higher than in the mouse-to-human direction for 18 of the 24 concepts. Input resistance, rheobase, membrane time constant and after-hyperpolarization transfer in both directions, whereas resting potential, latency and threshold voltages transfer poorly in both ($r<0.2$); on the same 24 concepts, mouse-to-human transfer gives a mean $r$ of 0.29 and a retention of 48\%. With the gene identities of the mouse input permuted, the mean $r$ is $-0.02$. An affine calibration on 20 paired mouse cells gives a held-out error of 1.03 times that of a calibration on 100 cells, and 67\% of the concepts are then predicted better than by a constant.

\section{Hyperparameters, parameter counts and training details}
\label{app:hyper}

\begin{table}[h]\centering\caption{Cohorts. In-domain benchmarks use five-fold cross-validation; zero-shot targets are scored through pre-registered feature crosswalks against measured electrophysiology.}\label{tab:datasets}
\small
\setlength{\tabcolsep}{2.5pt}
\begin{tabular}{p{1.05in} r r p{0.8in} p{0.85in} p{1.4in}}
\toprule
\raggedright Cohort & Cells & Genes & \raggedright Ephys features & \raggedright Labels & \raggedright Role in the paper \tabularnewline
\midrule
\raggedright Mouse visual cortex (Gouwens 2020) & 3,654 & 1,293 & \raggedright 39 & \raggedright 6 families, 4 layers & \raggedright In-domain benchmark; source cohort of every zero-shot transfer \tabularnewline
\raggedright Mouse motor cortex (Scala 2021) & 1,208 & 1,277 & \raggedright 29 & \raggedright 9 families, 4 layers & \raggedright In-domain benchmark; cross-area transfer, both directions \tabularnewline
\raggedright Human cortex (Lee 2023) & 612 / 704 & 1,293 & \raggedright 75 / 25 concepts & \raggedright 4 subclasses, 3 layers & \raggedright In-domain benchmark (612 complete cases); zero-shot cross-species target (704 paired cells, 247 acute) \tabularnewline
\raggedright Human L1 (Chartrand 2023) & 240 & 1,293 & \raggedright 26 concepts & \raggedright 2 subclasses (Lamp5/Pax6, Vip) & \raggedright Zero-shot target under composition shift \tabularnewline
\bottomrule
\end{tabular}

\par\smallskip
{\footnotesize\raggedright\noindent Cells = neurons with paired transcriptome and electrophysiology (Lee: in-domain complete cases / paired cells scored zero-shot); genes = panel size after matching to the frozen gene-text vocabulary; concepts = laboratory-independent feature correspondences used only for scoring zero-shot transfer.\par}
\end{table}

Table~\ref{tab:hyper} lists the settings of both training stages; the paragraphs below give the details and selection rules.

\paragraph{Stage 1: alignment.}
Context vectors are computed on the training cells of the fold only: four global statistics of the log expression (mean, standard deviation, detection rate, coefficient of variation) and, for every transcriptomic family and cortical layer present in the training set, the conditional mean and detection rate ($C=24$, 30 and 18 on the three cohorts). Training is full-batch AdamW (learning rate $10^{-4}$, weight decay $10^{-4}$, gradient-norm clipping at 1.0, 2{,}000 epochs, $\tau=0.07$, seed 0): the first 200 epochs are a warm-up in which only the auxiliary regression from the electrophysiology tower is active (weight 1.0); the contrastive term and the auxiliary regression from the cell tower (weight 0.2) are then ramped in linearly over 50 epochs. The checkpoint with the lowest validation InfoNCE loss is kept. The same configuration is used on all three cohorts; nonlinear projectors (MLP, residual MLP) did not improve validation InfoNCE in preliminary runs.

\begin{table}[h]\centering\caption{Hyperparameters and trainable parameter counts of the two stages.}\label{tab:hyper}
\small
\setlength{\tabcolsep}{4pt}
\begin{tabular}{l p{3.6in}}
\toprule
Component & Setting \tabularnewline
\midrule
\multicolumn{2}{l}{\textit{Stage 1: context adapter and alignment}} \tabularnewline
Context vector $C$ & 24 (visual), 30 (motor), 18 (human): 4 global + per-family and per-layer mean and detection rate \tabularnewline
Adapter & LN $\to$ Linear($C\to128$) $\to$ GELU $\to$ Linear($128\to3072$); gated residual, gate scale $s=0.1$ \tabularnewline
Projections $P_{\mathrm{cell}}, P_{\mathrm{ephys}}$ & LN $\circ$ Linear($3072\to3072$) $\circ$ LN \tabularnewline
Trainable parameters & adapter 0.40M; $P_{\mathrm{cell}}$ and $P_{\mathrm{ephys}}$ 9.45M each; auxiliary regression head 9.56M \tabularnewline
Contrastive loss & symmetric InfoNCE, temperature $\tau=0.07$ \tabularnewline
Regularizers & gene-cosine $\lambda_g=0.5$; cell-preserve $\lambda_c=0.1$; residual $\ell_2$ $10^{-4}$; auxiliary feature regression \tabularnewline
Optimizer & AdamW, lr $10^{-4}$, weight decay $10^{-4}$, full batch \tabularnewline
Schedule & 2{,}000 epochs; 200-epoch warm-up (auxiliary regression only) + 50-epoch ramp of the contrastive weight \tabularnewline
Model selection & lowest validation contrastive loss \tabularnewline
\multicolumn{2}{l}{\textit{Stage 2: forward heads (mouse)}} \tabularnewline
Head & per feature: LN(3072) $\to$ 512 $\to$ 128 $\to$ 1, dropout 0.2 (selected on the validation fold; default 2048 $\to$ 512, dropout 0.1) \tabularnewline
Optimizer & AdamW, lr $10^{-4}$, weight decay $10^{-3}$, $\le$1{,}500 epochs with early stopping \tabularnewline
Trainable parameters & 1.65M per feature (default 2048 $\to$ 512 head: 7.35M) \tabularnewline
Model selection & raw-scale MSE on the validation fold, per feature \tabularnewline
\multicolumn{2}{l}{\textit{Stage 2: forward heads (human)}} \tabularnewline
Head & per feature: LN(3072) $\to$ 256 $\to$ 1, dropout 0.3 \tabularnewline
Optimizer & AdamW, lr $10^{-3}$, weight decay $10^{-2}$, $\le$800 epochs with early stopping \tabularnewline
Trainable parameters & 0.79M per feature \tabularnewline
Selection of the configuration & validation Pearson among four candidates (Table~\ref{tab:human-indomain}) \tabularnewline
\multicolumn{2}{l}{\textit{Embedder}} \tabularnewline
Text embedder & text-embedding-3-large ($d=3072$), frozen; gene descriptions (GenePT) and verbalized ephys records embedded once and cached \tabularnewline
\bottomrule
\end{tabular}

\end{table}

\paragraph{Stage 2: forward heads.}
Each electrophysiological feature (39, 29 and 75 on the three cohorts) has its own MLP with layer normalization on the input, trained on the standardized target (log-transformed for heavy-tailed positive features) with AdamW and batch size 128; for each feature the epoch with the lowest validation raw-scale MSE is kept. The head width is chosen per cohort on the validation fold among four candidates: $2048$--$512$ (dropout 0.1, weight decay $10^{-5}$, learning rate $10^{-4}$, 4{,}000 epochs, no early stopping; the default), $1024$--$256$ (dropout 0.2, weight decay $10^{-4}$, $\le$2{,}000 epochs with early stopping), $512$--$128$ (dropout 0.2, weight decay $10^{-3}$, $\le$1{,}500 epochs) and $256$ (dropout 0.3, weight decay $10^{-2}$, learning rate $10^{-3}$, $\le$800 epochs). On the mouse cohorts $512$--$128$ is selected on both; on the human cohort $256$ is selected by nested validation (below). The transfer, attribution and interpretability analyses use the default $2048$--$512$ heads, trained before this selection. As a check of the selection rule on motor cortex, the four candidates scored with the inner-split protocol of the human cohort (heads trained on 85\% of the training cells, epochs selected on the remaining 15\%, scored on the validation fold) rank $1024$--$256$ ($0.505\pm0.027$), $2048$--$512$ ($0.502\pm0.023$), $512$--$128$ ($0.499\pm0.025$) and $256$ ($0.490\pm0.024$), all within one fold standard deviation of each other; on the test fold the two smaller heads score $0.520$ and $0.519$ against $0.509$ for the default, so the choice among them does not affect the conclusions.

\paragraph{Human forward head (nested validation).}
With 397 training and 92 validation cells per fold, the mouse head configuration overfits, and reusing the validation fold both to select the epoch per feature and to compare configurations makes validation scores optimistic (for the mouse configuration the validation mean is 0.470 while the test mean is 0.400). We therefore select the human head by nested validation: an inner split (15\% of the training cells) selects the epoch per feature, and the outer validation fold selects the configuration among the four candidates by its five-fold mean Pearson $r$ (0.394, 0.400, 0.408 and 0.410 for $2048$--$512$, $1024$--$256$, $512$--$128$ and $256$; Table~\ref{tab:human-indomain}); the test fold is untouched during selection. The selected 256-unit head scores $0.414\pm0.030$ on the test folds.

\paragraph{Baselines.}
Baselines that produce a cell embedding (Plain GenePT, Geneformer, UCE, Nicheformer) use one multi-output head with the default hyperparameters; we do not mix head types across methods. The fine-tuned Nicheformer baseline unfreezes the last transformer block (2.1M parameters) and trains it end-to-end with a 1024--512 multi-output head on the standardized targets (batch 32, learning rates $3\cdot10^{-4}$ for the head and $10^{-6}$ for the block, weight decay $10^{-5}$, at most 200 epochs, early stopping with patience 20 on the validation loss, best-validation weights); its inputs are the raw counts with mouse genes mapped to human Ensembl orthologs (16,463 and 15,131 genes in the Nicheformer vocabulary for visual and motor cortex). It overfits: at the selected epoch (12--27 of at most 200) the training loss is $0.51$--$0.72$ against a validation loss of $0.65$--$0.94$ (scaled MSE; $3.08$ on one human fold). Geneformer and Nicheformer embed the raw counts with mouse genes mapped to their human Ensembl orthologs (17,172 genes in the visual cohort; 15,131 for Nicheformer on motor cortex); UCE uses its own species-aware tokenization. On the human cohort, whose release contains CPM-normalized expression rather than raw counts, Geneformer, UCE and Nicheformer (frozen and fine-tuned) receive the CPM matrix in place of counts (30{,}669 Ensembl-mapped genes, 18{,}260 of them in the Nicheformer vocabulary), which differs from their pretraining input. The GenePT--ephys VAE and JAMIE keep their mouse configurations on every cohort; at that configuration the VAE does not train on motor or human cortex (its training loss stays at its initial value; Table~\ref{tab:forward}).

\paragraph{Per-feature comparison with the baselines.}
Table~\ref{tab:app-wins} lists, for every baseline of Table~\ref{tab:forward}, the number of features on which \method{} has the higher held-out Pearson $r$ or AUROC. Against the Plain GenePT MLP, the margin in $r$ is $+0.017$ on visual cortex, $+0.018$ on motor cortex and $+0.027$ on human cortex (paired Wilcoxon test over features: $p=1.8\times10^{-8}$, $2.3\times10^{-3}$ and $7\times10^{-5}$). On motor cortex the raw MSE is lower for the single-head baselines (691 for the Plain GenePT MLP, 695 and 687 for the frozen and the fine-tuned Nicheformer, 718 for \method{}); raw MAE and AUROC are best for \method{} on every cohort.

\begin{table}[h]\centering
\caption{Number of features on which \method{} is more accurate than each baseline of Table~\ref{tab:forward} (five-fold mean of the held-out Pearson $r$ or AUROC), with the $p$-value of a two-sided binomial test in parentheses.}
\label{tab:app-wins}
\small
\setlength{\tabcolsep}{3pt}
\resizebox{\textwidth}{!}{%
\begin{tabular}{l cc cc cc}
\toprule
 & \multicolumn{2}{c}{Mouse visual (of 39)} & \multicolumn{2}{c}{Mouse motor (of 29)} & \multicolumn{2}{c}{Human Lee (of 75)} \tabularnewline
\cmidrule(lr){2-3}\cmidrule(lr){4-5}\cmidrule(lr){6-7}
Baseline & $r$ & AUROC & $r$ & AUROC & $r$ & AUROC \tabularnewline
\midrule
Plain GenePT MLP & 36 ($4{\times}10^{-8}$) & 37 ($3{\times}10^{-9}$) & 22 ($0.008$) & 25 ($1{\times}10^{-4}$) & 48 ($0.020$) & 46 ($0.064$) \tabularnewline
Geneformer MLP & 36 ($4{\times}10^{-8}$) & 39 ($4{\times}10^{-12}$) & 29 ($4{\times}10^{-9}$) & 29 ($4{\times}10^{-9}$) & 49 ($0.011$) & 46 ($0.064$) \tabularnewline
UCE MLP & 36 ($4{\times}10^{-8}$) & 39 ($4{\times}10^{-12}$) & 23 ($0.002$) & 29 ($4{\times}10^{-9}$) & 70 ($1{\times}10^{-15}$) & 71 ($7{\times}10^{-17}$) \tabularnewline
Nicheformer MLP & 36 ($4{\times}10^{-8}$) & 39 ($4{\times}10^{-12}$) & 22 ($0.008$) & 25 ($1{\times}10^{-4}$) & 55 ($6{\times}10^{-5}$) & 57 ($7{\times}10^{-6}$) \tabularnewline
Fine-tuned Nicheformer & 36 ($4{\times}10^{-8}$) & 36 ($4{\times}10^{-8}$) & 21 ($0.024$) & 24 ($5{\times}10^{-4}$) & 58 ($2{\times}10^{-6}$) & 61 ($4{\times}10^{-8}$) \tabularnewline
GenePT--ephys VAE & 39 ($4{\times}10^{-12}$) & 39 ($4{\times}10^{-12}$) & 29 ($4{\times}10^{-9}$) & 29 ($4{\times}10^{-9}$) & 75 ($5{\times}10^{-23}$) & 73 ($2{\times}10^{-19}$) \tabularnewline
JAMIE & 36 ($4{\times}10^{-8}$) & 38 ($1{\times}10^{-10}$) & 27 ($2{\times}10^{-6}$) & 28 ($1{\times}10^{-7}$) & 43 ($0.25$) & 50 ($0.005$) \tabularnewline
\bottomrule
\end{tabular}
}
\end{table}

\section{Verbalization of electrophysiology}
\label{app:text}

\paragraph{Template.}
Starting from the cohort's feature table, protocol-index columns are removed, features with more than 40\% missing values are dropped, remaining gaps are median-imputed, and every feature is winsorized to its 1st--99th percentiles within the cohort. Values receive within-cohort quintile labels (\texttt{very\_low} to \texttt{very\_high}), and feature identifiers are mapped to readable names with units by a per-cohort feature-name map (e.g., \texttt{threshold\_t\_ramp} $\to$ ``Ramp spike threshold time [s]''). Each cell is rendered as ``Cell electrophysiology profile.'' followed by ``\emph{name}: \emph{winsorized value} (\emph{bin})'' for every feature in a fixed order and closed by four profile tokens that collapse designated features to low/mid/high: \emph{excitability} from the f--I slope, \emph{adaptation} from the adaptation index, \emph{spike\_shape} from the upstroke/downstroke ratio and \emph{membrane\_state} from the resting potential, each with a fallback feature when the primary one is absent. The rendering is deterministic and label-free. Records are embedded once with \texttt{text-embedding-3-large} (3072 dimensions) and L2-normalized; human records are rendered from the 75 Lee~2023 features with the same template and a human feature-name map, so no feature correspondence with mouse is needed to embed them. In the feature-name map, the mouse first-spike latency is labelled ``[ms]'' while its values are in seconds, and the records were embedded with this label.

\paragraph{Example (one mouse visual cortex cell, 39 features; abbreviated).}
\begin{quote}\small
``Cell electrophysiology profile. Resting membrane potential [mV]: $-80.4909$ (very\_low); Input resistance [MOhm]: 157.4218 (low); Sag ratio [unitless]: 0.0352 (low); Membrane time constant [ms]: 17.4863 (high); Membrane potential during sag step [mV]: $-98.5000$ (very\_low); F-I curve slope (excitability gain) [Hz/pA]: 0.3924 (high); Spike-frequency adaptation index [unitless]: 0.2096 (very\_high); First-spike latency [ms]: 0.0287 (mid); Average inter-spike interval [ms]: 120.8571 (very\_high); Long-square upstroke/downstroke ratio [unitless]: 2.1004 (mid); Long-square spike peak voltage [mV]: 6.7812 (very\_low); Long-square spike peak time [s]: 0.5824 (low); \dots; Short-square spike threshold voltage [mV]: $-54.6875$ (very\_low); Short-square spike threshold current [pA]: 560.0000 (high); Short-square spike threshold time [s]: 0.5030 (low). Profile tokens: excitability=high; adaptation=high; spike\_shape=mid; membrane\_state=low.'' %
\end{quote}

\section{Integration metrics and protocol}
\label{app:integration}

\begin{table}[h]\centering\caption{Integration metrics on held-out cells (FOSCTTM, label-transfer accuracy, silhouette) and aligner cost.}\label{tab:integration}
\small
\setlength{\tabcolsep}{3.5pt}
\resizebox{\textwidth}{!}{%
\begin{tabular}{l ccccc}
\toprule
Method & FOSCTTM$\downarrow$ & LTA family$\uparrow$ & LTA type$\uparrow$ & LTA layer$\uparrow$ & Silhouette$\uparrow$ \tabularnewline
\midrule
\multicolumn{6}{l}{\textit{Mouse visual}} \tabularnewline
Ours & \textbf{0.107 $\pm$ 0.004} & \textbf{0.926 $\pm$ 0.032} & \textbf{0.298 $\pm$ 0.020} & \textbf{0.462 $\pm$ 0.012} & \textbf{0.178 $\pm$ 0.040} \tabularnewline
JAMIE & 0.208 $\pm$ 0.006 & 0.777 $\pm$ 0.033 & 0.207 $\pm$ 0.022 & 0.407 $\pm$ 0.026 & 0.102 $\pm$ 0.020 \tabularnewline
GenePT--ephys VAE & 0.265 $\pm$ 0.136 & 0.629 $\pm$ 0.189 & 0.113 $\pm$ 0.064 & 0.301 $\pm$ 0.138 & 0.124 $\pm$ 0.175 \tabularnewline
Nicheformer + CLIP & 0.187 $\pm$ 0.004 & 0.850 $\pm$ 0.017 & 0.239 $\pm$ 0.028 & 0.434 $\pm$ 0.025 & 0.057 $\pm$ 0.007 \tabularnewline
\midrule
\multicolumn{6}{l}{\textit{Mouse motor}} \tabularnewline
Ours & \textbf{0.135 $\pm$ 0.010} & \textbf{0.848 $\pm$ 0.072} & \textbf{0.258 $\pm$ 0.028} & \textbf{0.492 $\pm$ 0.065} & \textbf{0.153 $\pm$ 0.027} \tabularnewline
JAMIE & 0.222 $\pm$ 0.013 & 0.551 $\pm$ 0.057 & 0.171 $\pm$ 0.021 & 0.412 $\pm$ 0.065 & 0.060 $\pm$ 0.029 \tabularnewline
GenePT--ephys VAE & 0.483 $\pm$ 0.012 & 0.145 $\pm$ 0.073 & 0.039 $\pm$ 0.028 & 0.397 $\pm$ 0.080 & -0.216 $\pm$ 0.037 \tabularnewline
Nicheformer + CLIP & 0.235 $\pm$ 0.026 & 0.637 $\pm$ 0.046 & 0.189 $\pm$ 0.032 & 0.462 $\pm$ 0.022 & 0.014 $\pm$ 0.018 \tabularnewline
\bottomrule
\end{tabular}
}
\par\medskip
\resizebox{\textwidth}{!}{%
\begin{tabular}{l r r r r l}
\toprule
Method & Epochs & Time (s) & GPU mem.\ (MB) & Params & Correspondence \tabularnewline
\midrule
Ours (stage 1) & 2,000 & 63 & 1,430 & 28.86M & none \tabularnewline
JAMIE (dual VAE) & $\approx$4,400 (early stop) & 200 & 362 & 2.17M & $O(n^2)$ cell pairs \tabularnewline
Fine-tuned Nicheformer (last block + head) & 40 (early stop) & $\approx$1,300$^{\dagger}$ & 18,348 & 3.17M & n/a (supervised) \tabularnewline
\bottomrule
\end{tabular}
}

\par\smallskip
{\footnotesize\raggedright\noindent FOSCTTM (lower is better): fraction of cells of the other modality that lie closer than the true partner; LTA: label-transfer accuracy across modalities for family, type and layer; silhouette on L2-normalized joint embeddings, averaged over the two modalities; best per column in bold. Only methods with a joint space are shown; JAMIE is scored on held-out cells with the same folds (its released notebook is transductive, fitting on all cells with full correspondence). Bottom: time to train each model to completion under its own schedule (the alignment stage for ours and JAMIE; the last transformer block and the multi-output head for the fine-tuned Nicheformer), and peak allocated GPU memory, on one NVIDIA H100 for one visual-cortex fold (2,374 training cells). $^{\dagger}$Extrapolated from the logged rate of a standalone run (16 epochs in 564 s) to the early-stopping epoch; the memory is measured with the same counter as the other rows.\par}
\end{table}

\paragraph{Definitions.}
\emph{FOSCTTM} (fraction of samples closer than the true match): for paired held-out cells, the fraction of opposite-modality cells that lie strictly closer (Euclidean) to a cell than its true partner, averaged over cells and both directions; 0 is a perfect pairing and 0.5 a random one. \emph{Label-transfer accuracy} (LTA): a $k$-nearest-neighbour classifier ($k=5$, JAMIE's default) is fit on the electrophysiology-side embeddings with their labels and predicts the labels of the transcriptomic-side embeddings of the same cells; reported for broad family, fine type and layer. \emph{Silhouette}: mean silhouette width of the broad-family labels within each modality's embedding, averaged over the two modalities (Euclidean distance on the L2-normalized embeddings). Each method is scored on its own embedding pair (dimensions are never mixed), cells are matched by identifier, the label vector is identical across methods, and values are aggregated as mean $\pm$ s.d.\ over the five test folds.

\paragraph{JAMIE.}
JAMIE's shared latent is computed by transforming the held-out test cells of each fold with the PCA, scaler and model fitted on that fold; the resulting FOSCTTM is $0.208\pm0.006$ (visual) and $0.222\pm0.013$ (motor). JAMIE's released notebook reports 0.002 on the same data because it fits and transforms all cells jointly with the full correspondence available (transductive), whereas our value is computed on held-out cells, so the two are not comparable.

\paragraph{Nicheformer+CLIP.}
This baseline replaces the gene side by frozen Nicheformer cell embeddings and aligns them with the text embeddings of the records through two linear projections (256 dimensions) trained with the symmetric InfoNCE loss (Adam, learning rate $10^{-3}$, batch size 256, 1{,}500 epochs). It reaches a FOSCTTM of 0.187 and 0.235 and a family LTA of 0.850 and 0.637 on visual and motor cortex. The contribution of the context adapter to the alignment of \method{} is measured in Appendix~\ref{app:ablations} by removing the adapter with all other settings fixed.

\paragraph{Time and memory.}
Table~\ref{tab:integration} (bottom) reports the time to train the alignment stage to completion for one fold (visual fold 0; 2{,}374 training cells) on one NVIDIA H100, each method in its own process under its own schedule, with the peak allocated GPU memory. Stage 1 (context adapter, two $3072\times3072$ affine projections and the auxiliary head, full-batch InfoNCE, 2{,}000 epochs) trains in 63\,s with 1{,}430\,MB; JAMIE's dual variational autoencoders (2.17M parameters and an $O(n^{2})$ cell-cell correspondence estimated inside \texttt{fit\_transform}; at most 10{,}000 epochs, early-stopped after about 4{,}400) take 200\,s and 362\,MB. Stage 1 is therefore about three times faster than JAMIE but uses about four times its GPU memory, because it trains full-batch on 3072-dimensional embeddings. The fine-tuned Nicheformer of Table~\ref{tab:forward} (last transformer block and a 1024--512 head, batch 32, early-stopped at epoch 40 in fold 0) peaks at 18{,}348\,MB, and its training time, extrapolated from the logged rate of a standalone run (16 epochs in 564\,s) to the early-stopping epoch, is about 1{,}300\,s, roughly 20 times stage 1 and 6 to 7 times JAMIE. Times are the CUDA-synchronized training loop of one complete run; data loading, the export of embeddings and the one-off embedding of gene descriptions and records are excluded for every method.

\section{Reverse readout details}
\label{app:reverse}

\begin{table}[h]\centering\caption{Reverse readouts on held-out cells (mean $\pm$ s.d.\ over five folds): macro-F1 for family, fine type and layer; Pearson $r$ for marker genes and programs.}\label{tab:reverse}
\small
\setlength{\tabcolsep}{3pt}
\resizebox{\textwidth}{!}{%
\begin{tabular}{l cccc cccc}
\toprule
 & \multicolumn{4}{c}{Mouse visual} & \multicolumn{4}{c}{Mouse motor} \tabularnewline
\cmidrule(lr){2-5}\cmidrule(lr){6-9}
Method & Family F1 & Type F1 & Layer F1 & Marker $r$ & Family F1 & Type F1 & Layer F1 & Marker $r$ \tabularnewline
\midrule
Ours (context-guided multitask head) & \textbf{0.779 $\pm$ 0.028} & 0.382 $\pm$ 0.017 & \textbf{0.459 $\pm$ 0.019} & \textbf{0.539 $\pm$ 0.007} & \textbf{0.710 $\pm$ 0.075} & \textbf{0.572 $\pm$ 0.043} & \textbf{0.575 $\pm$ 0.035} & \textbf{0.572 $\pm$ 0.009} \tabularnewline
Direct supervised ephys & 0.755 $\pm$ 0.018 & \textbf{0.397 $\pm$ 0.019} & 0.413 $\pm$ 0.017 & 0.453 $\pm$ 0.031 & 0.678 $\pm$ 0.047 & 0.557 $\pm$ 0.083 & 0.486 $\pm$ 0.020 & 0.509 $\pm$ 0.014 \tabularnewline
\midrule
Ephys $\to$ GenePT PCA-ridge & 0.665 $\pm$ 0.028 & 0.183 $\pm$ 0.031 & 0.404 $\pm$ 0.016 & 0.456 $\pm$ 0.032 & 0.625 $\pm$ 0.034 & 0.380 $\pm$ 0.055 & 0.477 $\pm$ 0.040 & 0.509 $\pm$ 0.015 \tabularnewline
Ephys $\to$ Geneformer MLP & 0.745 $\pm$ 0.023 & 0.320 $\pm$ 0.026 & 0.429 $\pm$ 0.034 & 0.450 $\pm$ 0.111 & 0.363 $\pm$ 0.220 & 0.210 $\pm$ 0.154 & 0.313 $\pm$ 0.102 & 0.216 $\pm$ 0.154 \tabularnewline
Ephys $\to$ UCE MLP & 0.659 $\pm$ 0.042 & 0.150 $\pm$ 0.007 & 0.344 $\pm$ 0.043 & 0.391 $\pm$ 0.021 & 0.672 $\pm$ 0.065 & 0.383 $\pm$ 0.042 & 0.496 $\pm$ 0.036 & 0.463 $\pm$ 0.030 \tabularnewline
Ephys $\to$ Nicheformer MLP & 0.751 $\pm$ 0.034 & 0.280 $\pm$ 0.050 & 0.366 $\pm$ 0.049 & 0.488 $\pm$ 0.046 & 0.665 $\pm$ 0.057 & 0.411 $\pm$ 0.081 & 0.469 $\pm$ 0.089 & 0.521 $\pm$ 0.019 \tabularnewline
JAMIE reverse & 0.549 $\pm$ 0.013 & 0.083 $\pm$ 0.003 & 0.362 $\pm$ 0.017 & 0.403 $\pm$ 0.005 & 0.336 $\pm$ 0.030 & 0.142 $\pm$ 0.008 & 0.307 $\pm$ 0.031 & 0.383 $\pm$ 0.010 \tabularnewline
\bottomrule
\end{tabular}
}
\end{table}

\paragraph{Targets.}
Transcriptomic family (6 classes on visual cortex, 9 on motor cortex), fine transcriptomic type (60 and 76 labels; a class is eligible in a fold if it has at least 10 training and 3 test cells, giving 49.2 and 21.2 eligible labels per fold on average), cortical layer (6 and 4 classes), the log-normalized expression of 32 marker genes, and 9 marker programs defined as the mean log expression of their genes.
Markers: \emph{Pvalb, Sst, Vip, Lamp5, Sncg, Gad1, Gad2, Slc6a1, Slc17a7, Slc17a6, Fezf2, Bcl11b, Tle4, Foxp2, Rorb, Cux2, Reln, Npy, Nos1, Tac1, Htr3a, Kcnc1, Kcnip2, Cacna1e, Cacna2d3, Hcn1, Scn1a, Kcnq2, Calb1, Calb2, Penk, Crh}. Programs: GABAergic (Gad1, Gad2, Slc6a1); glutamatergic (Slc17a7, Slc17a6); Pvalb fast-spiking (Pvalb, Kcnc1); Sst (Sst, Npy); Vip (Vip, Htr3a); Lamp5/Reln (Lamp5, Reln); deep excitatory (Bcl11b, Tle4, Foxp2); upper-layer IT (Cux2, Rorb); ion channel (Kcnc1, Kcnip2, Cacna1e, Cacna2d3, Hcn1, Scn1a, Kcnq2). Markers absent from a panel are skipped (27 of the 32 are present and evaluated).

\paragraph{Readout.}
(1)~A regressor maps the standardized feature vector $\vy_i$ to the aligned transcriptomic embedding $\vz^{\mathrm{cell}}_i$. It is an MLP trained with a regression loss on the embedding, $0.45\,\mathrm{MSE}+0.45\,(1-\cos)$, and with three terms computed through frozen models: the forward heads applied to the predicted embedding recover the measured features (weight 0.2), and logistic readouts of family, type and layer (weight 0.05) and ridge readouts of marker genes and programs (weights 0.03 and 0.05), fit on the true embeddings of the training cells, are applied to the predicted embedding (AdamW, learning rate $3\cdot10^{-4}$, weight decay $10^{-5}$, at most 400 epochs with patience 70 on the validation loss). (2)~The predicted embedding is standardized and reduced to its leading 64 principal components (fit on training cells) and concatenated with $\vy_i$. (3)~One multitask head, two hidden layers of 256 and 128 units with layer normalization and dropout 0.25, ends in three softmax heads (family, type, layer) and two linear regression heads (markers, programs); it is trained with a balanced sum of the task losses (AdamW, learning rate $10^{-3}$, weight decay $10^{-4}$, batch 128, at most 200 epochs with patience 30, gradient clipping 5). No stage-1 parameter is updated. The \emph{direct supervised} control applies logistic regression ($C=1$, class-weight balanced) and ridge regression ($\alpha=10$) to the standardized feature vector without any embedding. Classification is scored by macro-F1 and regression by Pearson $r$ on held-out cells; this readout is above the per-task supervised control on 9 of 10 endpoints (4/5 visual, 5/5 motor) and 1.4--2.1$\times$ JAMIE on family F1 (Table~\ref{tab:reverse}).

\section{Crosswalks, permutation nulls and calibration protocol}
\label{app:crosswalk}

\paragraph{Mouse visual $\leftrightarrow$ motor (11 concepts).}
The two laboratories extract different feature sets (39 Allen IPFX features; 29 features of \citealp{scala2021phenotypic}) with no shared names. Before any cross-cohort score was computed we fixed 11 biophysical concepts with one feature (or a derived difference) on each side and an expected sign; three concepts (sag, AHP, adaptation) were registered as sign-uncertain because the two pipelines define them with possibly opposite conventions.
\begin{table}[h]
\centering
\small
\caption{Pre-registered visual$\leftrightarrow$motor crosswalk.}
\label{tab:app-crosswalk-mouse}
\resizebox{\textwidth}{!}{%
\begin{tabular}{lllc}
\toprule
Concept & Visual feature (IPFX) & Motor feature (Scala) & Sign certain \\
\midrule
resting potential & \texttt{vrest} & \texttt{Resting.membrane.potential..mV.} & yes \\
input resistance & \texttt{ri} & \texttt{Input.resistance..MOhm.} & yes \\
membrane time constant & \texttt{tau} & \texttt{Membrane.time.constant..ms.} & yes \\
sag & \texttt{sag} & \texttt{Sag.ratio} & no \\
rheobase & \texttt{threshold\_i\_long\_square} & \texttt{Rheobase..pA.} & yes \\
AP threshold & \texttt{threshold\_v\_long\_square} & \texttt{AP.threshold..mV.} & yes \\
AP amplitude & \texttt{peak\_v} $-$ \texttt{threshold\_v} (long square) & \texttt{AP.amplitude..mV.} & yes \\
upstroke/downstroke & \texttt{upstroke\_downstroke\_ratio\_long\_square} & \texttt{Upstroke.to.downstroke.ratio} & yes \\
AHP & \texttt{fast\_trough\_v} $-$ \texttt{threshold\_v} (long square) & \texttt{Afterhyperpolarization..mV.} & no \\
latency & \texttt{latency} & \texttt{Latency..ms.} & yes \\
adaptation & \texttt{adaptation} & \texttt{Spike.frequency.adaptation} & no \\
\bottomrule
\end{tabular}}
\end{table}

\paragraph{Mouse $\to$ human (25 Lee, 26 Chartrand concepts).}
The matching rules were fixed before any human score was computed: mouse \texttt{*\_long\_square} first-spike waveform features pair with the human rheobase-sweep (\texttt{\_rheo}) columns, the spike-train features \texttt{adaptation} and \texttt{avg\_isi} with the \texttt{\_hero} sweep, and \texttt{latency} with \texttt{latency\_rheo}; where a human table gives only threshold-relative voltages (Lee) the mouse side uses the derived difference, and where absolute voltages exist (Chartrand) absolute pairs are used; time landmarks, spike widths and chirp features have no counterpart and are excluded without counting as failures; all expected signs are positive. The 25 Lee concepts cover passive properties (resting potential, input resistance, time constant, sag and the sag-step potential), excitability (rheobase, f--I slope, latency, mean ISI, adaptation and the ramp threshold current) and first-spike waveform (upstroke/downstroke ratio, threshold, amplitude, fast AHP and threshold-relative trough voltage) on the long-square, ramp and short-square sweeps; the Chartrand crosswalk shares 22 of them, uses absolute trough voltages and adds the absolute long-square peak and fast-trough voltages (26 concepts).

\paragraph{Scoring, null and retention.}
For each concept the signed Pearson $r$ between the fold-mean prediction and the measured value across target cells is computed per fold and averaged. The permutation null shuffles the cell correspondence $N=1000$ times (seed 0) and records the 95th percentile of $|r|$ (null median $r\approx0$; 95th percentiles 0.056 and 0.033 for visual$\to$motor and motor$\to$visual); a concept passes if its $r$ exceeds this percentile with the pre-registered sign. Retention is $R=\overline{|r_{\mathrm{target}}|}/\overline{|r_{\mathrm{in\text{-}domain}}|}$ over the same concepts, where the in-domain reference is the held-out (out-of-fold) score of the target cohort's own model when one exists (mouse$\leftrightarrow$mouse) and the source model's own held-out score otherwise (human targets, pre-registered). Verdicts: A (transfer holds) if $R\ge0.5$ and at least two thirds of the concepts pass the null with the correct sign; B (partial) if $0.3\le R<0.5$, or $R\ge0.5$ without the null condition; C (failure) if $R<0.3$, reported with the collapse magnitude and diagnostics. Method comparisons use a two-sided concept-paired Wilcoxon test, reported whether or not significant.

\paragraph{Few-shot calibration.}
Because pipelines differ in liquid-junction correction, temperature, solutions and protocols, predictions are on the source pipeline's scale. For each concept and each $n\in\{5,10,20,50,100\}$ we draw $n$ target cells without replacement, fit an affine map $\hat y=a\,x+b$ on them, evaluate the mean absolute error on the remaining cells, and take the median over 200 resamples; the reference is the MAD of a constant predictor, $\mathrm{mean}|y-\mathrm{median}(y)|$. The pre-registered criterion for ``about 20 paired cells suffice'' is a concept-median ratio $\mathrm{MAE}(20)/\mathrm{MAE}(100)\le1.05$ and at least 60\% of concepts with $\mathrm{MAE}(20)$ below the MAD. On the Lee cohort the ratio is 1.025 (met) and 56\% of concepts (14/25) beat the MAD (not met); on its acute-slice subset the ratio is 1.027 and 60\% beat the MAD (both met). Calibration cannot rescue concepts that did not transfer (resting potential, latency, adaptation, threshold voltages), and on Chartrand's L1 cells only 4\% of concepts beat the MAD.

\section{Human in-domain benchmark (Lee 2023)}
\label{app:human-indomain}

\begin{table}[h]\centering\caption{Human-Lee in-domain benchmark (612 cells, 75 features) and the nested-validation head selection.}\label{tab:human-indomain}
\small
\setlength{\tabcolsep}{3pt}
\resizebox{\textwidth}{!}{%
\begin{tabular}{l c c c c c c c c}
\toprule
Method & Pearson $r$ & AUROC & scaled MSE & raw MAE & $\Delta$ vs.\ GenePT & feat.\ better & $p$ (feat.) & $p$ (fold) \tabularnewline
\midrule
Ours (head selected on validation) & \textbf{0.414 $\pm$ 0.030} & \textbf{0.725} & 1.182 & 9.82 & +0.027 & 48/75 & $7.2\times10^{-5}$ & 0.062 \tabularnewline
Ours (mouse head config., no selection) & 0.400 $\pm$ 0.024 & 0.722 & 1.221 & 10.00 & +0.013 & 44/75 & 0.051 & 0.31 \tabularnewline
Mouse stage-1 frozen + human heads & 0.395 $\pm$ 0.032 & 0.722 & 1.225 & 10.03 & +0.008 & 36/75 & 0.55 & 0.62 \tabularnewline
Plain GenePT MLP & 0.387 $\pm$ 0.021 & 0.713 & 1.196 & 9.90 & ref. & \textemdash & \textemdash & \textemdash \tabularnewline
Geneformer MLP & 0.401 $\pm$ 0.034 & 0.719 & 1.206 & 9.88 & +0.014 & 42/75 & 0.13 & 0.31 \tabularnewline
UCE MLP & 0.312 $\pm$ 0.035 & 0.670 & 1.265 & 10.65 & -0.075 & 11/75 & $1.7\times10^{-10}$ & 0.062 \tabularnewline
Nicheformer MLP & 0.386 $\pm$ 0.028 & 0.710 & 1.214 & 10.13 & -0.001 & 29/75 & 0.21 & 1.00 \tabularnewline
Fine-tuned Nicheformer & 0.386 $\pm$ 0.023 & 0.710 & 1.207 & 10.06 & -0.001 & 28/75 & 0.15 & 0.81 \tabularnewline
GenePT--ephys VAE & 0.001 $\pm$ 0.018 & 0.510 & 1.385 & 11.82 & -0.371 & 3/75 & $3.3\times10^{-13}$ & 0.062 \tabularnewline
JAMIE & 0.406 $\pm$ 0.034 & 0.718 & 0.824 & 10.03 & +0.019 & 32/75 & 0.46 & 0.19 \tabularnewline
\bottomrule
\end{tabular}
}
\par\medskip
\begin{tabular}{l p{2.6in} c c}
\toprule
Head & Configuration & Validation $r$ & Test $r$ \tabularnewline
\midrule
tiny & 256, dropout 0.3, wd $10^{-2}$, lr $10^{-3}$, $\le$800 epochs, early stop & \textbf{0.410 $\pm$ 0.018} & 0.414 $\pm$ 0.030 \tabularnewline
small & 512--128, dropout 0.2, wd $10^{-3}$, $\le$1500 epochs, early stop & 0.408 $\pm$ 0.013 & 0.417 $\pm$ 0.030 \tabularnewline
mid & 1024--256, dropout 0.2, wd $10^{-4}$, $\le$2000 epochs, early stop & 0.400 $\pm$ 0.018 & 0.411 $\pm$ 0.027 \tabularnewline
prod & 2048--512, dropout 0.1, wd $10^{-5}$, 4000 epochs, no early stop (mouse config.) & 0.394 $\pm$ 0.015 & 0.400 $\pm$ 0.024 \tabularnewline
\bottomrule
\end{tabular}

\par\smallskip
{\footnotesize\raggedright\noindent Top: Human Lee 2023 in-domain benchmark (612 cells, 75 features, subclass-stratified 5-fold): 5-fold test mean $\pm$ s.d. of the feature-averaged Pearson $r$, AUROC, scaled MSE and raw MAE; per-feature $\Delta$ vs.\ Plain GenePT MLP, number of features improved, and paired Wilcoxon $p$ over features and over folds. Bottom: forward-head configurations compared on the validation fold (5-fold mean $\pm$ s.d.); the best validation score selects the 256-unit head used as ``Ours'' in the main text; the last column gives the test scores.\par}
\end{table}

\paragraph{Design.}
From the 704 GABAergic neurons of \citet{lee2023signature} we keep the 75 IPFX features with at most 10\% missing values and then the 612 cells with no missing value among them (complete-case design), with the 1{,}293-gene panel of the mouse visual cohort. Five folds are stratified by subclass with 397/92/123 train/validation/test cells per fold. Stage 1 is trained from scratch on the human pairs with the mouse configuration (Appendix~\ref{app:hyper}); the forward head is selected by nested validation because 92 validation cells are too few to select both the epoch per feature and the configuration (Appendix~\ref{app:hyper}). Verbalized records use the human feature-name map (Appendix~\ref{app:text}); the context dimension is 18.

\paragraph{Results.}
With the selected head the proposed representation scores $0.414\pm0.030$, against $0.406\pm0.034$ for JAMIE (43/75 features better in $r$, $p=0.25$; 50/75 in AUROC, $p=5\times10^{-3}$), $0.401\pm0.034$ for Geneformer ($p=0.016$; 49/75), $0.387\pm0.021$ for Plain GenePT ($p=7\cdot10^{-5}$; 48/75), $0.386\pm0.028$ and $0.386\pm0.023$ for the frozen and the fine-tuned Nicheformer (55/75 and 58/75) and $0.312\pm0.035$ for UCE (70/75); the GenePT--ephys VAE does not train at its mouse configuration ($r=0.001$). With the mouse head configuration the proposed method scores $0.400\pm0.024$, and freezing the mouse stage 1 and training only human heads gives $0.395\pm0.032$. The margin over JAMIE is not significant in $r$; this is not a claim of human state of the art.

\paragraph{Low-resource curve.}
With a ridge head trained on $n$ labelled human cells (same folds; ten random draws per fold; medians over the 50 runs), the mouse-frozen representation (mouse stage 1, no human electrophysiology seen before the head) is the best representation at $n\le50$.
\begin{table}[h]
\centering
\small
\caption{Low-resource human forward prediction: mean Pearson $r$ over 75 features with a ridge head trained on $n$ labelled Lee~2023 cells (median over 5 folds $\times$ 10 draws; ``all'' uses the full training fold, 5 folds). The last row's stage 1 saw the electrophysiology of all training cells and is a reference, not a low-resource result.}
\label{tab:app-lowresource}
\begin{tabular}{lccccc}
\toprule
Representation (+ ridge head) & $n=25$ & $n=50$ & $n=100$ & $n=200$ & all \\
\midrule
Mouse stage 1, frozen & 0.270 & 0.297 & 0.329 & 0.367 & 0.412 \\
Plain GenePT & 0.197 & 0.264 & 0.333 & 0.374 & 0.420 \\
Geneformer & 0.195 & 0.256 & 0.305 & 0.347 & 0.374 \\
\midrule
Human stage 1 (saw all pairs; reference only) & 0.306 & 0.338 & 0.369 & 0.400 & 0.431 \\
\bottomrule
\end{tabular}
\end{table}

\section{Attribution}
\label{app:attribution}

\begin{figure}[h]\centering\includegraphics[width=\linewidth]{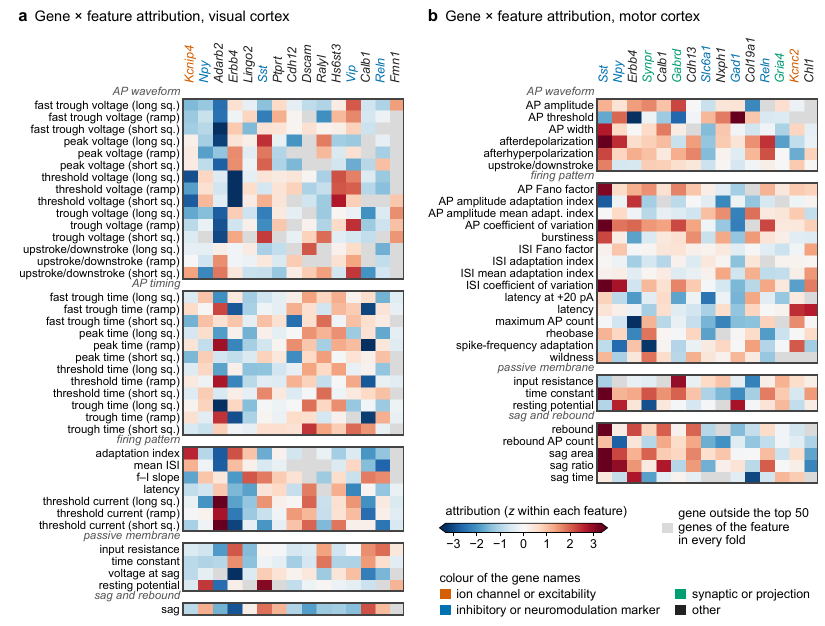}\caption{Gene $\times$ feature attribution of the forward heads in (a) visual and (b) motor cortex (gradient$\times$input, mean over five folds, $z$-scored per feature) for the fifteen most attributed genes (columns), with the features (rows) grouped by physiological family; grey cells, gene outside the top 50 genes of that feature in every fold.}\label{fig:interp}\end{figure}

\paragraph{Method.}
Three models with a differentiable path from the log expression of the panel to every predicted feature are attributed with one protocol on the same held-out cells and folds: the proposed model (context-adapted gene pooling, aligned cell embedding and the default per-feature heads, $2048$--$512$), the Plain GenePT MLP (frozen pooled embedding and its own head) and JAMIE, whose gene$\to$electrophysiology imputation is re-implemented as a differentiable path (fitted 512-component PCA and standardization, gene encoder, latent mean, electrophysiology decoder, inverse transform; the path reproduces JAMIE's predictions on every fold). Attributions are gradient$\times$input difference, $\partial\hat y_f/\partial\vx\odot(\vx-\vx_0)$, with integrated gradients~\citep{sundararajan2017axiomatic} (8 steps) as corroboration; the baseline $\vx_0$ is the mean expression profile of the training fold; rankings are restricted to the shared gene panel (JAMIE's input has nine extra genes). A gene's global score is its mean absolute attribution over cells and features. The attribution explains the default-head models; the validation-selected heads of Table~\ref{tab:forward} differ little from them in accuracy (Appendix~\ref{app:hyper}).

\emph{Readouts.} \emph{Marker recovery}: fraction of the train-fold cell-type marker union (25 genes per label by mean difference on training cells, so no held-out information enters) among the top-$K$ genes, $K\le400$; null: a random ranking (hypergeometric mean and 95th percentile). \emph{Gene-set enrichment}: fraction of set members among the top-50 genes divided by their fraction in the panel, for curated inhibitory-interneuron markers, neuromodulation/neuropeptide genes, synaptic-transmission genes and an ion-channel/excitability set defined by a rule on the symbol (every panel gene beginning with \emph{Scn}, \emph{Kcn}, \emph{Hcn} or \emph{Cacn} except the lncRNA \emph{Kcnq1ot1}; 40 genes including auxiliary and interacting subunits such as \emph{Kcnip4}); the same rule colours the ion-channel gene labels in Fig.~\ref{fig:interp}; null: random gene sets of equal size (hypergeometric 95th percentile). Ion-channel genes are the canonical transcriptomic correlates of intrinsic electrophysiology~\citep{tripathy2017transcriptomic,bomkamp2019transcriptomic}. \emph{Stability}: mean pairwise Jaccard overlap of the top-50 sets across folds (random: 0.02) and mean pairwise Spearman correlation of the full ranking (null: permuted rankings). \emph{Deletion and keep-only}: for every feature, the top-$k$ genes are reset to the baseline profile (or every gene except the top-$k$ is), predictions are recomputed and the change of held-out Pearson $r$ is compared with twenty random $k$-gene sets. A single gene is one part in about 1{,}300 of the pooled cell vector, so occluding a single gene has a negligible effect for every model, and genes are therefore deleted in sets. \emph{Joint space}: following JAMIE's Fig.~6, genes are reset to the train mean in the transcriptome input, the transcriptome embeddings are recomputed with the electrophysiology embeddings fixed, and the change of the broad-family label-transfer accuracy (Sec.~\ref{sec:results-reverse}) is recorded for every single gene and for each model's own top-$k$ genes against random sets; the unperturbed value reproduces the reported accuracy on every fold. JAMIE's notebook ablated its all-cell integration model with a classifier fitted and tested on the gene latent alone; here both models are the fold models of the paper and the score is cross-modal.

\begin{table}[h]\centering\caption{Attribution benchmark: the proposed model, the Plain GenePT MLP and JAMIE, attributed with the same gradient$\times$input protocol on the same held-out cells and folds, scored on biological plausibility, stability across folds, faithfulness under gene deletion and keep-only, and gene removal in the joint space.}\label{tab:app-attribution}
\footnotesize
\setlength{\tabcolsep}{3pt}
\resizebox{\textwidth}{!}{%
\begin{tabular}{l c c c c}
\toprule
\multicolumn{5}{l}{\emph{Mouse visual cortex}} \tabularnewline
\midrule
Criterion & Ours & Plain GenePT MLP & JAMIE & null \tabularnewline
\midrule
Train-fold markers: recovery AUC ($K\le400$) & \textbf{0.57 $\pm$ 0.01} & 0.57 $\pm$ 0.02 & 0.06 $\pm$ 0.03 & 0.16$^{\mu}$ \tabularnewline
Train-fold markers among the top 200 (fraction) & 0.64 $\pm$ 0.02 & \textbf{0.64 $\pm$ 0.03} & 0.05 $\pm$ 0.03 & 0.21 \tabularnewline
Curated inhibitory markers among the top 200 (fraction of 11) & 0.40 $\pm$ 0.05 (4.4/11) & \textbf{0.60 $\pm$ 0.05 (6.6/11)} & 0.00 $\pm$ 0.00 (0.0/11) & 0.36 \tabularnewline
Train-fold markers: fold enrichment, top 50 & \textbf{6.60 $\pm$ 0.34} & 6.60 $\pm$ 0.44 & 0.35 $\pm$ 0.33 & 1.78 \tabularnewline
Inhibitory markers: fold enrichment, top 50 & \textbf{9.40 $\pm$ 0.00} & 8.93 $\pm$ 1.97 & 0.00 $\pm$ 0.00 & 4.70 \tabularnewline
Ion-channel/excitability genes (rule set): fold enrichment, top 50 & \textbf{2.84 $\pm$ 0.98} & 1.94 $\pm$ 0.46 & 1.29 $\pm$ 0.91 & 2.59 \tabularnewline
Neuromodulation/neuropeptide genes: fold enrichment, top 50 & 14.65 $\pm$ 2.36 & \textbf{16.38 $\pm$ 1.93} & 0.00 $\pm$ 0.00 & 4.31 \tabularnewline
Synaptic-transmission genes: fold enrichment, top 50 & 2.59 $\pm$ 0.89 & \textbf{3.23 $\pm$ 1.14} & 0.32 $\pm$ 0.72 & 3.23 \tabularnewline
\midrule
Stability across folds: Jaccard of top-50 sets & 0.55 $\pm$ 0.06 & \textbf{0.55 $\pm$ 0.04} & 0.13 $\pm$ 0.04 & 0.02$^{\mu}$ \tabularnewline
Stability across folds: Spearman $\rho$ of full rankings & \textbf{0.88 $\pm$ 0.02} & 0.82 $\pm$ 0.01 & 0.47 $\pm$ 0.02 & 0.05 \tabularnewline
\midrule
Deletion of the top-10 genes per feature: $\Delta r$ minus random & 0.005 $\pm$ 0.001 [34/39] & \textbf{0.010 $\pm$ 0.002 [34/39]} & 0.006 $\pm$ 0.003 [26/39] & 0.000$^{\mu}$ \tabularnewline
Deletion of the top-50 genes per feature: $\Delta r$ minus random & 0.022 $\pm$ 0.003 [36/39] & \textbf{0.039 $\pm$ 0.006 [37/39]} & 0.012 $\pm$ 0.004 [27/39] & 0.000$^{\mu}$ \tabularnewline
Keep-only top-50 genes per feature: held-out $r$ & \textbf{0.43 $\pm$ 0.01} & 0.42 $\pm$ 0.02 & 0.24 $\pm$ 0.02 & \textemdash \tabularnewline
Keep-only 50 random genes per feature: held-out $r$ & 0.23 $\pm$ 0.02 & 0.18 $\pm$ 0.02 & 0.14 $\pm$ 0.03 & \textemdash \tabularnewline
\midrule
Joint space: LTA change, own top-50 genes removed & -0.018 $\pm$ 0.012 (random +0.000) & \textemdash & -0.007 $\pm$ 0.011 (random -0.006) & \textemdash \tabularnewline
Gradient$\times$input vs.\ integrated gradients (rank $\rho$) & 0.99 $\pm$ 0.00 & 0.99 $\pm$ 0.00 & 0.94 $\pm$ 0.01 & \textemdash \tabularnewline
\bottomrule
\end{tabular}
}
\par\medskip
\resizebox{\textwidth}{!}{%
\begin{tabular}{l c c c c}
\toprule
\multicolumn{5}{l}{\emph{Mouse motor cortex}} \tabularnewline
\midrule
Criterion & Ours & Plain GenePT MLP & JAMIE & null \tabularnewline
\midrule
Train-fold markers: recovery AUC ($K\le400$) & \textbf{0.47 $\pm$ 0.03} & 0.34 $\pm$ 0.02 & 0.12 $\pm$ 0.01 & 0.16$^{\mu}$ \tabularnewline
Train-fold markers among the top 200 (fraction) & \textbf{0.51 $\pm$ 0.03} & 0.37 $\pm$ 0.02 & 0.12 $\pm$ 0.02 & 0.20 \tabularnewline
Curated inhibitory markers among the top 200 (fraction of 11) & 0.80 $\pm$ 0.04 (8.8/11) & \textbf{0.82 $\pm$ 0.00 (9.0/11)} & 0.15 $\pm$ 0.05 (1.6/11) & 0.36 \tabularnewline
Train-fold markers: fold enrichment, top 50 & \textbf{5.74 $\pm$ 0.45} & 3.98 $\pm$ 0.16 & 0.54 $\pm$ 0.30 & 1.59 \tabularnewline
Inhibitory markers: fold enrichment, top 50 & \textbf{16.72 $\pm$ 1.94} & 14.86 $\pm$ 1.27 & 0.46 $\pm$ 1.04 & 4.64 \tabularnewline
Ion-channel/excitability genes (rule set): fold enrichment, top 50 & \textbf{2.94 $\pm$ 0.73} & 0.38 $\pm$ 0.57 & 0.89 $\pm$ 0.57 & 2.55 \tabularnewline
Neuromodulation/neuropeptide genes: fold enrichment, top 50 & \textbf{15.32 $\pm$ 2.33} & 13.62 $\pm$ 1.90 & 0.00 $\pm$ 0.00 & 4.26 \tabularnewline
Synaptic-transmission genes: fold enrichment, top 50 & 5.11 $\pm$ 0.71 & \textbf{9.26 $\pm$ 2.62} & 1.92 $\pm$ 0.71 & 3.19 \tabularnewline
\midrule
Stability across folds: Jaccard of top-50 sets & \textbf{0.52 $\pm$ 0.05} & 0.43 $\pm$ 0.05 & 0.11 $\pm$ 0.02 & 0.02$^{\mu}$ \tabularnewline
Stability across folds: Spearman $\rho$ of full rankings & \textbf{0.88 $\pm$ 0.01} & 0.76 $\pm$ 0.03 & 0.37 $\pm$ 0.02 & 0.06 \tabularnewline
\midrule
Deletion of the top-10 genes per feature: $\Delta r$ minus random & 0.011 $\pm$ 0.004 [22/29] & \textbf{0.016 $\pm$ 0.005 [25/29]} & 0.005 $\pm$ 0.005 [17/29] & 0.000$^{\mu}$ \tabularnewline
Deletion of the top-50 genes per feature: $\Delta r$ minus random & 0.029 $\pm$ 0.013 [25/29] & \textbf{0.044 $\pm$ 0.015 [26/29]} & 0.014 $\pm$ 0.013 [21/29] & 0.000$^{\mu}$ \tabularnewline
Keep-only top-50 genes per feature: held-out $r$ & 0.42 $\pm$ 0.03 & \textbf{0.44 $\pm$ 0.02} & 0.19 $\pm$ 0.02 & \textemdash \tabularnewline
Keep-only 50 random genes per feature: held-out $r$ & 0.23 $\pm$ 0.03 & 0.19 $\pm$ 0.01 & 0.11 $\pm$ 0.01 & \textemdash \tabularnewline
\midrule
Joint space: LTA change, own top-50 genes removed & -0.046 $\pm$ 0.016 (random -0.004) & \textemdash & -0.019 $\pm$ 0.042 (random -0.008) & \textemdash \tabularnewline
Gradient$\times$input vs.\ integrated gradients (rank $\rho$) & 0.98 $\pm$ 0.00 & 0.99 $\pm$ 0.00 & 0.97 $\pm$ 0.01 & \textemdash \tabularnewline
\bottomrule
\end{tabular}
}
\par\smallskip
{\footnotesize\raggedright\noindent Five-fold mean $\pm$ s.d.\ on held-out cells; bold marks the best of the three models where a direction exists. Nulls: recovery AUC and Jaccard, mean of a random ranking ($^{\mu}$); recall, enrichment and Spearman, 95th percentile of a random ranking or of random gene sets of equal size (hypergeometric); deletion, zero by construction (random sets of the same size are subtracted). Deletion rows show in brackets the number of features on which the top genes hurt more than random genes. The joint-space row is the change of the broad-family label-transfer accuracy after removing each model's own top-50 genes (random-set change in parentheses); the Plain GenePT MLP has no joint space. Attribution describes reliance of the trained model and is hypothesis-generating, not causal.\par}
\end{table}

\paragraph{Figures~\ref{fig:reverse}a,b and~\ref{fig:interp}.}
The UMAP panels of Fig.~\ref{fig:reverse} use the held-out cells of fold 0 (731 visual, 242 motor): the $\ell_2$-normalized cell and electrophysiology embeddings of the aligned space are concatenated, one UMAP (n\_neighbors 200, min\_dist 0.5, seed 42) is fit on the union and applied to each modality, following the latent-space displays of JAMIE; grey segments join a random subset of pairs, and families with fewer than ten held-out cells are drawn in grey. The gene$\times$feature maps of Fig.~\ref{fig:interp} show, for the fifteen genes with the best mean attribution rank across features, the signed gradient$\times$input averaged over the five folds and $z$-scored within each feature; features are grouped by physiological family (action-potential waveform, action-potential timing, firing pattern, passive membrane, sag and rebound) and gene names are coloured by gene class (ion channel/excitability, inhibitory or neuromodulation marker, synaptic or projection, other). In the maps of visual cortex, \emph{Erbb4} and \emph{Adarb2} carry the spike-shape features and \emph{Sst}, \emph{Calb1} and \emph{Erbb4} the sag features. The marker-recovery curves, gene-removal curves and the remaining benchmark criteria are tabulated in Table~\ref{tab:app-attribution}.

\paragraph{Results.}
JAMIE's imputation model, attributed with the same protocol, prioritizes genes with little biological structure: its marker recovery is at or below the random-ranking level, its top-50 sets overlap by 0.11 to 0.13 across folds, and removing its top genes changes its joint space about as much as random genes. Between the two language-prior models, the proposed model recovers markers faster and is more enriched for markers and ion-channel genes on motor cortex, ties the Plain GenePT MLP on visual cortex, and has the more stable full ranking on both cohorts; the Plain GenePT MLP is more enriched for synaptic-transmission genes (visual and motor) and for neuromodulation genes on visual cortex, and its predictions lose more accuracy when its own top genes are deleted (its reliance is more concentrated, whereas keeping only random genes retains more accuracy for the proposed model). Only the proposed model and JAMIE have a joint space; the proposed model's top genes are the ones its joint space depends on. None of this establishes that a gene causes a feature.

\section{Full transfer tables}
\label{app:transfer-tables}

\begin{table}[h]\centering\caption{Zero-shot transfer summary across cortical areas and species: mean $r$, concepts above the permutation null with the registered sign, retention and verdict.}\label{tab:transfer}
\small
\setlength{\tabcolsep}{2.5pt}
\begin{tabular}{l c l c c c c}
\toprule
Target & cells / concepts & Method & mean $r$ & $>$ null & $R$ (verdict) & $p$ \tabularnewline
\midrule
Visual $\to$ motor & 1,208 / 11 & Ours & 0.344 & 11/11 & 0.70 (A) & 0.17 \tabularnewline
 &  & Plain GenePT & 0.317 & 11/11 & 0.66 (A) & ref. \tabularnewline
\midrule
Motor $\to$ visual & 3,654 / 11 & Ours & 0.384 & 11/11 & 0.71 (A) & 0.46 \tabularnewline
 &  & Plain GenePT & 0.371 & 11/11 & 0.69 (A) & ref. \tabularnewline
\midrule
Lee 2023, all & 704 / 25 & Ours & 0.282 & 18/25 & 0.47 (B) & 0.10 \tabularnewline
 &  & Plain GenePT & 0.271 & 18/25 & 0.44 (B) & ref. \tabularnewline
 &  & Gene-permuted (ctrl.) & -0.001 & 8/25 & 0.16 (C) &  \tabularnewline
\midrule
Lee 2023, acute & 247 / 25 & Ours & 0.335 & 20/25 & 0.58 (A) & 0.075 \tabularnewline
 &  & Plain GenePT & 0.320 & 18/25 & 0.55 (A) & ref. \tabularnewline
\midrule
Chartrand L1 & 240 / 26 & Ours & 0.077 & 6/26 & 0.17 (C) & 0.47 \tabularnewline
 &  & Plain GenePT & 0.082 & 7/26 & 0.16 (C) & ref. \tabularnewline
\midrule
Mouse L1 anchor$^\dagger$ & 515 / 26 & Ours & 0.497 & 26/26 & 0.85 (A) & 0.0086 \tabularnewline
 &  & Plain GenePT & 0.457 & 26/26 & 0.78 (A) & ref. \tabularnewline
\bottomrule
\end{tabular}

\par\smallskip
{\footnotesize\raggedright\noindent Per concept, Pearson $r$ between the 5-fold-mean prediction and the measured value (feature correspondences fixed before scoring); $>$ null: concepts whose $r$ exceeds the 95th percentile of a 1,000-permutation null with the registered sign; retention $R$ = mean $|r|$ divided by the mouse in-domain held-out mean $|r|$ over the same concepts; verdict (pre-registered): A if $R\ge0.5$ and $\ge2/3$ of concepts above null, B if $0.3\le R<0.5$, C otherwise. $p$: two-sided paired Wilcoxon over concepts, Ours vs.\ Plain GenePT. $^\dagger$Cells the visual model was trained on (476/547 overlap): an upper reference, not a score.\par}
\end{table}

Table~\ref{tab:transfer} gives the per-target summary for every method; the per-concept values are in the released tables.

\paragraph{Mouse visual $\leftrightarrow$ motor.}
Without retraining, the visual model applied to the 1{,}208 motor cells has mean $|r|$ 0.401 against an in-domain reference of 0.573 (retention 70\%) and the motor model applied to the 3{,}654 visual cells 0.442 against 0.624 (71\%); all 11 concepts exceed the permutation null in both directions (signed means 0.344 and 0.384). Plain GenePT reaches 0.317 and 0.371 (differences to ours $+0.027$, $p=0.175$ and $+0.013$, $p=0.465$; not significant), UCE 0.270 and 0.346, Geneformer 0.052 and 0.040, and Nicheformer 0.326 and 0.385 ($p=0.365$ and $p=1.0$ against ours).
Adaptation transfers with $r\approx-0.32$ in both directions and for every method because the two pipelines define the index with opposite orientation; the concept was registered as sign-uncertain before scoring, so the magnitude counts and the sign is a convention. Resting potential transfers worst (17--35\% retention), as expected for the quantity most sensitive to recording conditions. Raw-unit errors consist mostly of fixed pipeline offsets (98\% of the rheobase, 99\% of the AP-amplitude and 100\% of the sag error), which is why cross-cohort accuracy is scored by correlation and offsets are handled by the few-shot calibration of Appendix~\ref{app:crosswalk}. An affine calibration fit on $n$ paired target cells gives a held-out error of 1.165, 1.076, 1.036, 1.014, 1.007 and 1.005 times that of a calibration fit on all target cells for $n=5$, 10, 20, 50, 100 and 200 (median over concepts and both directions).

\paragraph{Mouse $\to$ human.}
On the full Lee cohort (704 cells, 25 concepts) the proposed model scores mean $r$ 0.282 with 18/25 concepts above the null and $R=0.47$ (verdict B); on the 247 acute-slice cells 0.335, 20/25 and $R=0.58$ (A); on the cultured cells 0.271 and $R=0.46$ (B). Plain GenePT scores 0.271 and $R=0.44$ ($\Delta=+0.011$, $p=0.10$; not significant); a gene-permuted input scores $-0.001$ ($R=0.16$, C), with 8 of 25 concepts tied to the fast-spiking phenotype remaining at $r=0.09$--$0.18$, so about $0.1$ of the human correlation reflects gene-agnostic statistics of the expression vector. On Chartrand's 240 L1 cells (Lamp5 and Vip only) the model scores 0.077, 6/26 and $R=0.17$ (C), and 0.22 against a class-matched mouse reference (also C); the same mouse L1 cells scored through Chartrand's feature extraction agree with ours at a median $r$ of 0.996 over the 26 concepts, so the low transfer comes from the cell composition, not from the crosswalk. A mouse L1 anchor on cells seen in training gives $R=0.85$. Removing subclass means from both predictions and measurements lowers the Lee mean $r$ from 0.30 to 0.17, while the four subclass means correlate at 0.59 and the direction of the Pvalb offset is correct for 22/25 concepts: more than half of what transfers is between-subclass structure, but within-subclass signal remains.